\documentclass[review,number,sort&compress,12pt]{elsarticle}%

\usepackage{natbib}
\usepackage{geometry}
\usepackage{graphicx}
\usepackage{subfigure}
\usepackage{scrtime}
\usepackage{epstopdf}
\usepackage{enumerate}
\usepackage{amsmath}
\usepackage{siunitx}
\usepackage{amssymb}
\usepackage{upgreek}
\usepackage{appendix}
\usepackage[colorlinks, linkcolor=black, citecolor=black, anchorcolor=black]{hyperref}
\hypersetup{hidelinks}
\usepackage{algorithm}% http://ctan.org/pkg/algorithms
\usepackage{algorithmicx}
\usepackage{algpseudocode}% http://ctan.org/pkg/algorithmicx

\usepackage{multirow}
\usepackage{lineno}
\usepackage{enumerate}
\usepackage[section]{placeins}

\usepackage{booktabs}
\usepackage{threeparttable}

\usepackage{booktabs}
\usepackage[dvipsnames]{xcolor}
\usepackage{tcolorbox}
\usepackage{soul}
\begin{document}
% \modulolinenumbers[1]
% \linenumbers

\begin{frontmatter}
\title{Feature-adjacent multi-fidelity physics-informed machine learning for partial differential equations}

\author[1]{Wenqian Chen}
% \ead{xxx@pnnl.gov}

\author[1]{Panos Stinis \corref{cor1}}
\ead{panos.stinis@pnnl.gov}

% \author[1]{...}
% \ead{...}

\cortext[cor1]{Corresponding author}

\address[1]{Advanced Computing, Mathematics and Data Division \\
Pacific Northwest National Laboratory \\ Richland, WA 99354, USA}

\begin{abstract}
Physics-informed neural networks  have emerged as an alternative method for solving partial differential equations. However, for complex problems, the training of such networks can still require   high-fidelity data which can be expensive to generate. To reduce or even eliminate the dependency on high-fidelity data, we propose a novel multi-fidelity architecture which is based on a feature space shared by the low- and high-fidelity solutions. In the feature space, the projections of the low-fidelity and high-fidelity solutions are adjacent by constraining  their relative distance. The feature space is represented with an encoder and its mapping to the original solution space is effected through a decoder.  The proposed multi-fidelity approach is validated on forward and inverse problems for steady and unsteady problems described by partial differential equations.  

\end{abstract}


\end{frontmatter}

\section{Introduction}
\label{sec_intro}
Benefiting from the rapid development of computational capacity, optimization algorithms and automatic differentiation capabilities \cite{paszke2019pytorch, abadi2016tensorflow},  
physics-informed neural networks (PINNs)  have emerged as a powerful tool for solving forward and inverse  problems for partial differential equations (PDEs). First proposed in \cite{raissi2019physics}, the PINN approach tries to constrain the output of a neural network by means of PDEs and high-fidelity data (if any are available). When using only physics, the PINN approach has been shown to be effective for problems with reasonable smoothness and simple boundary conditions \cite{raissi_wang_triantafyllou_karniadakis_2019, liu2022hierarchical, yazdani2020systems}. However, it can face difficulties  for problems exhibiting sharp spatial features  and/or short time scales \cite{wang2021understanding,wight2020solving,mcclenny2020self}. 
In such cases, data can be required to further constrain the optimization of the neural network parameters. Often, high-fidelity data sets obtained from either measurements or numerical simulations can be very expensive to obtain (sometimes prohibitively so), while  much cheaper low-fidelity data can be abundant. Although they are less accurate, low-fidelity data are capable of providing useful information to direct a PINN towards better solutions. Therefore, by combining low- and high-fidelity information (LF and MF , respectively), multi-fidelity (MF) machine learning has the potential to greatly enhance our ability for solving PDEs.

In multi-fidelity machine learning, it is crucial to discover the relation between the low- and high-fidelity data (solutions). The relevant work in the literature can be broadly classified into three categories. The first approach uses neural networks to explicitly approximate the correlation between the low- and high-fidelity solutions.
This approach has been successfully applied to function approximation\cite{meng2020composite,howard2022multifidelity,mahmoudabadbozchelou2021data,guo2022multi, chen2022multi}, forward problems \cite{howard2022multifidelity, liu2019multi, ramezankhani2022data}, inverse problems \cite{meng2020composite}, and uncertainty quantification \cite{de2022bi, meng2021multi}.
The second approach is based on transfer learning, where part of a neural network is shared by the low- and high-fidelity solutions. The neural network is first trained with low-fidelity information,  and then trained with  high-fidelity information while keeping fixed the common part shared by the low- and high-fidelity solutions. This approach has been successfully applied to function approximation \cite{song2022transfer, chakraborty2021transfer, ashouri2022transfer, de2022neural, jiang2023use, li2022line}, forward problems \cite{chakraborty2021transfer, aliakbari2022predicting}, and uncertainty quantification \cite{de2020transfer}.
The third approach predicts the low- and high-fidelity solutions with the same network. This is effective when the low-fidelity solution is close to the high-fidelity solution in solution space. This approach has been successfully applied to  forward problems \cite{chen2021physics} and inverse problems \cite{basir2022physics}.

In the present work, we propose a novel approach to discover the relation between the low- and high-fidelity solutions. We embed the relation between the low- and high-fidelity solutions into a neural network by means of constraining their relative distance in a feature function space. Specifically, the low- and high-fidelity solutions share the same set of basis functions for the feature space, and the same mapping from the feature space to the original solution space.  The difference between the low- and high-fidelity solutions is reflected in their locations in the feature space, which is represented by a small number of trainable parameters. The low-fidelity  and high-fidelity outputs are trained together with a  composite loss function comprising of low- and high-fidelity information (physics or data). The high-fidelity output is not only constrained by the prior high-fidelity information but also restricted to be adjacent to the low-fidelity solution in the feature space. Our numerical experiments show that this approach helps the training algorithm to obtain a better approximation, and improve the  accuracy for the high-fidelity prediction.

The rest of the paper is structured as follows. Section \ref{sec_MFMethod} introduces the main idea and neural network architecture of our multi-fidelity approach. In Section \ref{sec_MFTrain}, we describe the construction of the loss function and some details of the training. Section \ref{sec_R_D} contains numerical results for steady and unsteady problems. Section \ref{sec_Discussion} offers a discussion of the results.

\section{Feature-adjacent multi-fidelity neural network architecture}
\label{sec_MFMethod}
\subsection{Feature space and feature-adjacency}
The key point in multi-fidelity modeling is how to represent the relation between the low- and high-fidelity solutions. Note that the low- and high- fidelity solutions are often similar to each other. Thus it is important to embed the similarity into the multi-fidelity  architecture. For this purpose, we propose three assumptions:
\begin{enumerate}[(1)]
\item Both the low- and high-fidelity solutions can be mapped to a  function space spanned by a set of basis functions $\mathbf{f}=\{f_i(\mathbf{x})\}_{i=1}^{N_f}$, where their projections $\mathbf{f}_L=\{\alpha_i^L f_i(\mathbf{x})\}_{i=1}^{N_f}$ and $\mathbf{f}_H=\{\alpha_i^H f_i(\mathbf{x})\}_{i=1}^{N_f}$ on this space are close. Here $\mathbf{x}\in \mathbb{R}^{n_p}$ is the input of the low- and high-fidelity solutions. 
\item The closeness relation is enforced by constraining the relative distance between the projection coefficients $\pmb{\alpha}^L=\{\alpha_i^L\}_{i=1}^{N_f}$ and $\pmb{\alpha}^H=\{\alpha_i^H\}_{i=1}^{N_f}$ of the low-and high-fidelity solutions, namely
\begin{equation}
	\label{eq_dis_LH}
	\|
	\frac{\pmb{\alpha}^H-\pmb{\alpha}^L}
	{\pmb{\alpha}^L}
	\|_\infty
	\le d_f
	,
\end{equation}
where $d_f$ is the relative distance. Eq. \eqref{eq_dis_LH} can also be transferred to an equality constraint
\begin{equation}
\label{eq_dis_LH2}	
\pmb{\alpha}^H=\pmb{\alpha}^L\otimes(1+d_f\pmb{\lambda}),
\end{equation}
where $\pmb{\lambda}\in[-1,1]^{N_f}$ and $\otimes$ is a point-wise multiplication operator. 
\item The projection coefficient of the low-fidelity solution is simply set as $ \pmb{\alpha}^L =\{1\}_{i=1}^{N_f}$.
\end{enumerate}

Based on these assumptions, we can write the relation between the low- and high-fidelity projections as follows:
\begin{equation}
	\label{eq_projection_LH}
	\begin{aligned}
	\mathbf{f}_L&=\mathbf{f} \\
	\mathbf{f}_H&=\mathbf{f}_L\otimes(1+d_f\pmb{\lambda}) \\
	\end{aligned}
\end{equation}

\begin{figure}[!ht]
	\centering
	\includegraphics{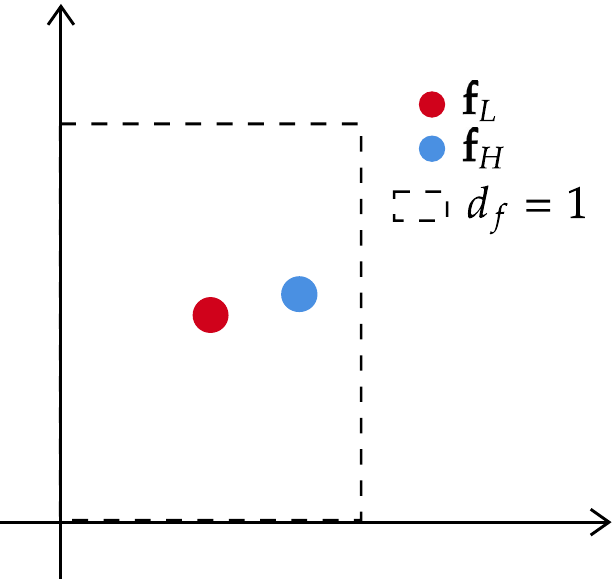}
	\caption{  A schematic of the projections of the low- and high-fidelity solutions in a two-dimensional feature-adjacent space.}
	\label{fig_features}
\end{figure}

The distribution of the low- and high-fidelity projection coefficients is illustrated in  Fig. \ref{fig_features}, where the high-fidelity projection coefficients are constrained to be inside the dash-lined box controlled by $d_f$.
In this work, we refer to the function basis as ``features", the function space as ``FA space (feature-adjacent space)" and $\pmb{\lambda}$ as ``feature shift". We further comment that the function basis can also be interpreted as a reduced basis in the sense of reduced-order modeling \cite{lucia2004reduced, maday2006reduced}. In this way, the similarity between the low- and high-fidelity solutions comes from the same basis, and their discrepancy results from their different coefficients.

Based on the above assumptions, we still need to find the FA space, build the mapping from the input to the FA space, and also build the mapping from the FA space to solution space. For general problems, there is usually no prior knowledge about the appropriate FA space. 

Deep neural networks have emerged as good candidates to address such questions, if adequate data or physics is available. Thus, we have opted to use a neural network to build the mapping from the input to the FA space, and another network to build the mapping from the FA space to the solution space. We call the first network  ``encoder net", while the second network  ``decoder net". 
The vector $\pmb{\lambda}$ is trainable, and is determined along with the encoder and decoder network training. According to
Eq. \eqref{eq_dis_LH2}, $\pmb{\lambda}$ should be constrained to $[-1,1]^{N_f}$. In our numerical experiments, we find that an explicit constraint is not necessary during the training process, provided that $\pmb{\lambda}$ is initialized within the range $[-1,1]^{N_f}$. This is because during the network training process,  $\pmb{\lambda}$ does not move far away from its initialization.
The relative distance $d_f$ can be determined based on prior knowledge. With a large $d_f$, the high-fidelity features can be far from the low-fidelity features. When $d_f \to 0$, the low- and high-fidelity solutions are very close.

%\vspace{-0.5cm}

\begin{figure}[!ht]
	\centering
	\includegraphics{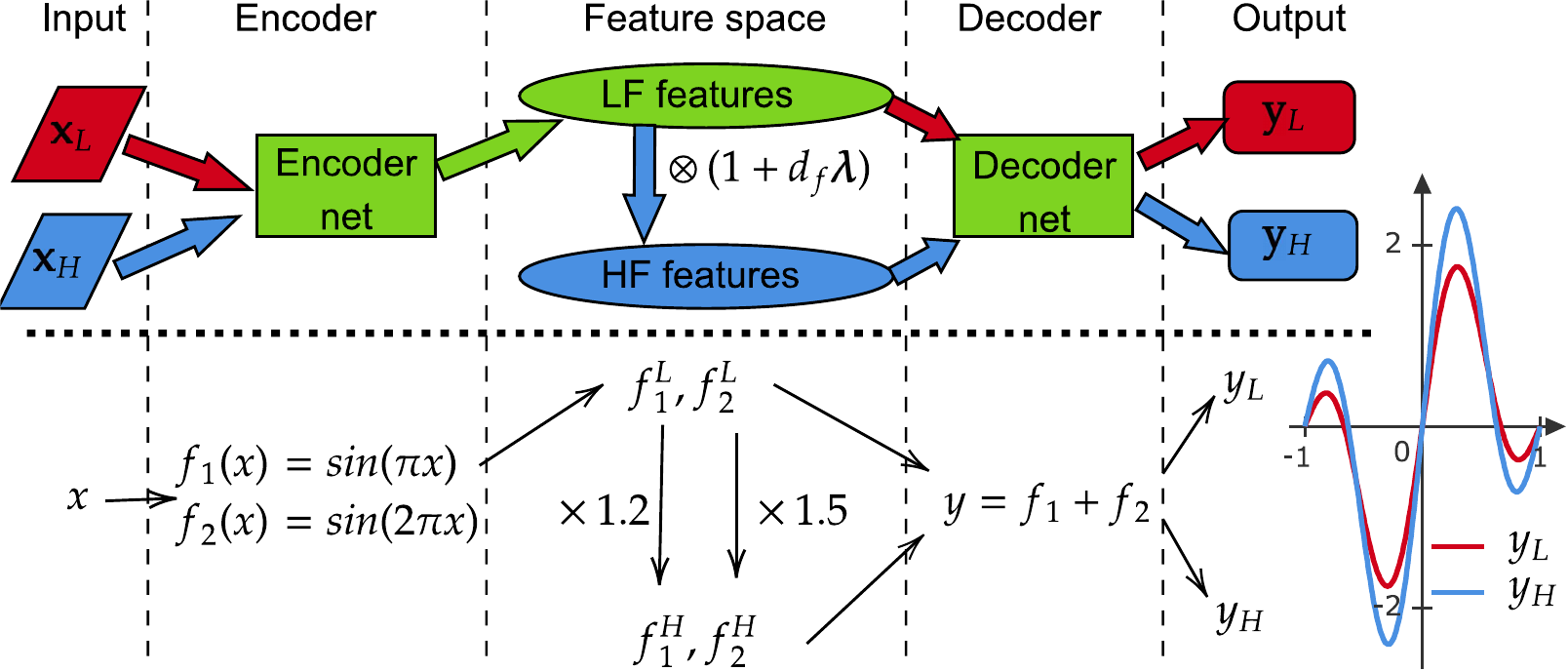}
	\caption{  (top) A schematic of the multi-fidelity network architecture and (bottom) a trivial example. In the top part, the green components are shared by the low- and high-fidelity propagation. The red and blue components are used for the low- and high-fidelity propagation, respectively. }
	\label{fig_flowChart}
\end{figure}

Putting together the ideas above, the multi-fidelity network architecture is depicted in the top of Fig. \ref{fig_flowChart}. We refer to the forward propagation of the encoder net and the decoder net as 
\begin{equation}
\begin{aligned}
\mathbf{f}&:=g_{e}(\mathbf{x}), \qquad && \text{Encoder net} \\
\mathbf{y}&:=g_{d}(\mathbf{f}), \qquad && \text{Decoder net} \\
\end{aligned}
.
\label{eq_en_decoder}
\end{equation}
Combining Eqs. \eqref{eq_projection_LH} and \eqref{eq_en_decoder},  the forward propagation needed to estimate the low- and high-fidelity outputs are
\begin{equation}
	\label{eq_forwardPropagation}
	\begin{aligned}
	\mathbf{y}_L&:=g_L(\mathbf{x})=g_d(\mathbf{f}_L)= g_d(\mathbf{f})=
	              g_{d}\left(  g_{e}( \mathbf{x} )  \right)  \\
	\mathbf{y}_H&:=g_H(\mathbf{x})=g_d(\mathbf{f}_H)=
                  g_{d}\left(  g_{e}(\mathbf{x}) \otimes \left(1+ d_f \pmb{\lambda}\right) \right)  \\
    \end{aligned}	
.        
\end{equation} 
Fig. \ref{fig_flowChart} shows an illustration of the architecture along with a trivial example.
%\vspace{-0.5cm}

\subsection{Encoder and decoder networks}
The networks employed for the encoder and decoder do not require a specific architecture, they can be chosen from standard network architectures, such as feedforward neural network (FNN), convolutional neural network (CNN) or recurrent neural network (RNN) etc. In the current work,  we focus on solving forward and inverse problems for PDEs and we have chosen to use FNNs as encoder/decoder. 

A FNN contains an input layer, $L$ hidden layers and an output layer. The $l_\text{th}$ layer has $n_l$ neurons, where $0 \le l \le L+1$ denotes the input layer, the $L$ hidden layers and the output layer (with a slight abuse of terminology, we also call the input and output also layers so that we have a more compact definition for the whole network). The standard forward propagation of the FNN, i.e. the function $\mathbf{y}=\mathcal{N}_{nn0}(\mathbf {x})$, is defined as follows:
\begin{equation}
	\left\{
	\begin{aligned}
		\mathbf{y}^{0} &= \mathbf{x} \\
		\mathbf{y}^{l} &= f_{act}( \mathbf{W}^{l}\mathbf{y}^{l-1}+ \mathbf{b}^{l}), \qquad  1\le l \le L \\
		\mathbf{y}  \;     &=\mathbf{y}^{L+1}=\mathbf{W}^{L+1}\mathbf{y}^{L}+ \mathbf{b}^{L+1} \\
	\end{aligned}
	\right .
	,
\end{equation}
where $\mathbf{y}^{l}  \in \mathbb{R}^{n_{l}} $ is the output of the $l_\text{th}$ layer, $\mathbf{W}^{l} \in \mathbb{R}^{n_{l}} \times \mathbb{R}^{n_{l-1}}$ are the weights, $\mathbf{b}^{l} \in \mathbb{R}^{n_{l}}$ are the biases, and $f_{act}$ is a point-wise activation function.
In \cite{wang2021understanding}, Wang et al. proposed an attention-based network architecture, whose performance is superior to the traditional FNN. Here we refer to this network architecture as ``modified FNN". The  forward propagation of the modified FNN, i.e. the function $\mathbf{y}={\mathcal{N}_{nn1}}(\mathbf {x})$, is defined as follows:
\begin{equation}
	\left\{
	\begin{aligned}
		\mathbf{U}\;&= f_{act}(\mathbf{W}^U\mathbf{x}+\mathbf{b}^U) &&\\
		\mathbf{V}\;&= f_{act}(\mathbf{W}^V\mathbf{x}+\mathbf{b}^V) &&\\
		\mathbf{y}^{1} &= f_{act}( \mathbf{W}^{1}\mathbf{x}+ \mathbf{b}^{1}) &&\\
		\mathbf{Z}^{l} &= f_{act}( \mathbf{W}^{l}\mathbf{y}^{l-1}+ \mathbf{b}^{l}), && 2\le l \le L \\
		\mathbf{y}^{l} &= (1-\mathbf{Z}^{l}) \otimes \mathbf{U} + \mathbf{Z}^{l} \otimes \mathbf{V}, && 2\le l \le L \\
		\mathbf{y} \;     &=\mathbf{y}^{L+1}=\mathbf{W}^{L+1}\mathbf{y}^{L}+ \mathbf{b}^{L+1} &&\\
	\end{aligned}
	\right .
	,
\end{equation}
where  $\{\mathbf{W}^U, \mathbf{W}^V\} \subset \mathbb{R}^{n_{1}} \times \mathbb{R}^{n_{0}}$ and $\{\mathbf{b}^{U}, \mathbf{b}^{V}\} \subset \mathbb{R}^{n_{1}}$ are the additional weights and biases compared to the traditional FNN. Note that there should be at least 2 hidden layers and all hidden layers should have a same number of neurons in the modified FNN, namely $n_1=n_2=...=n_L$ and $L \ge 2$.
In our numerical experiments, if the number of hidden layers $L \ge 2$ and all hidden layers have the same number of neurons, then employment of the modified FNN is implied, otherwise the traditional FNN is  employed.

To build the encoder net,  the FA space and the decoder net within the MF architecture, we employ $L_{M}$ layers of neurons in addition to the input and output layers.
We split the $L_{M}$ layers of neurons into three parts: the first $L_f-1$ layers, the $L_f$th layer (referred to as ``feature layer") and the last $L_{M}-L_f$ layers. Here we refer to $L_f$ as ``feature depth", it satisfies $0\le L_f \le L_M$. The encoder net is built with the input layer, the first $L_f-1$ layers as hidden layers, and the $L_f$th layer as the output layer. The decoder net is built with the $L_f$th layer as the input layer,  the last $L_{M}-L_f$ layers as the hidden layers and the output layer. The input layer size of the encoder net  and the output layer size of the decoder net are determined by the problem to be solved and data pre-processing. If $L_f=L_M$, the decoder net only contains the feature layer and the output layer, implying the decoder is only a linear mapping. If $L_f=0$,  the encoder is neglected, and the feature layer is equal to the input layer.
Also, if not stated otherwise, we choose the Swish function $f_{act}(x)=x\cdot \text{sigmoid}(x)$ \cite{ramachandran2017searching} as activation function.

\subsection{Normalization of input and output}

The training performance of neural networks is usually affected by the magnitude scale of the input/output \cite{loffe2014accelerating}, thus normalization before  the training process can help. 
For a specific problem, the minimum and maximum bound of the input $\mathbf{x} \in \mathbb{R}^{n_{p}}$ is always predefined, and thus can be used to scale the input $\mathbf{x}$ to lie in $[-1, 1]^{n_p}$:
\begin{equation}
	\begin{aligned}
	\widetilde{\mathbf{x}} &= s_{I}\left( \mathbf{x} \right)
	= \frac{\mathbf{x}-(\mathbf{x}_{\max}+\mathbf{x}_{\min})/2}{(\mathbf{x}_{\max} -\mathbf{x}_{\min})/2}		
	\end{aligned}
\label{eq_norm_input}
\end{equation}

For the output, a predefined range is not always available. If there is not any prior knowledge about the output, the normalization can only be the identical mapping, namely
\begin{equation}
	\widetilde{\mathbf{y}} = s_{O}\left( \mathbf{y} \right) =\mathbf{y}.
	\label{eq_norm_output1}	
\end{equation}
This is widely employed for PINN training which uses only physics.

On the other hand, when we have some prior knowledge about the output, such as a low- or high-fidelity input-output data set, we use the ``standard score" concept from statistics \cite{singh2020investigating}. To this end, the mean $\overline{\mathbf{y}}$ and standard derivation $\pmb{\sigma}_{\mathbf{y}}$ of the outputs are first calculated from the data set, and then they are used to  normalize the output:
\begin{equation}
	\widetilde{\mathbf{y}} = s_{O}\left( \mathbf{y} \right)= \frac{\mathbf{y}-\overline{\mathbf{y}}}{\pmb{\sigma}_{\mathbf{y}}}.
	\label{eq_norm_output2}	
\end{equation}

\subsection{Fourier feature embedding}
According to \cite{tancik2020fourier}, a simple transformation of the network input to a set of Fourier features can help the network training avoid the "spectral bias" problem \cite{basri2020frequency,rahaman2019spectral,wang2021eigenvector}. We will  describe briefly the construction of Fourier features from the input (see \cite{tancik2020fourier, wang2021eigenvector} for more details about Fourier feature embedding).

Fourier features are built by a random Fourier mapping $\gamma: \mathbf{x} \to \mathbb{R}^{2m}$:
\begin{equation}
	\label{eq_featureTransform}
	\gamma(\mathbf{x}) =
	 \left( 
	 \sin(\pi \mathbf{B} \mathbf{x}),
	 \cos(\pi \mathbf{B} \mathbf{x})
	\right), 
\end{equation}
where $\mathbf{B} \in \mathbb{R}^{m \times n_p}$ is a random matrix and $m$ is the number of Fourier features.  According to \cite{tancik2020fourier}, based on numerical experiments, sampling the elements of $\mathbf{B}$ from a Gaussian distribution is a quasi-optimal choice. In the current work, the $i$th column of $\mathbf{B}$ is sampled with a Gaussian distribution  $b_i\sim\mathcal{N}(0,\frac{\pi}{2}\sigma_i^2)$ for $i=1,2,...,n_p$. For the $i$th dimension, the frequency of each Fourier feature is equal to $\left|b_i\right|/2$.
The expectation of the number of waves $K_i$ for the $i$th dimension is  $\mathbb{E} K_i=\mathbb{E}\left|b_i\right|=\sigma_i$, given that the input $\mathbf{x}$ will be normalized to $[-1, 1]^{n_p}$ before Fourier feature embedding. Therefore, $\sigma_i$  can be interpreted as the average wave number along the $i$th dimension for $i=1,2,...,n_p$. Note that the average wave number $\pmb{\sigma}=\{\sigma_i\}_{i=1}^{n_p}$ and $m$ are user-defined hyperparameters. Therefore, there is no trainable parameter in this transformation, and thus it  can be deemed as data pre-processing.

\subsection{Summary}
\label{subsec_MF_summary}
Combining the input normalization \eqref{eq_norm_input} with the Fourier feature transformation \eqref{eq_featureTransform}, the input is transformed to Fourier features:
\begin{equation}
	\widehat{\mathbf{x}} = \gamma\left( s_{I}(\mathbf{x})\right), 
\end{equation}
where $\widehat{\mathbf{x}} \in \mathbb{R}^{2m}$.  The Fourier features are used as input for the MF network.
The forward propagation of the MF network can be built by   connecting all the components, namely the normalization of the input in Eq. \eqref{eq_norm_input}, the Fourier feature embedding in Eq. \eqref{eq_featureTransform}, the encoder/decoder net in Eq. \eqref{eq_forwardPropagation} and the output normalization in Eq. \eqref{eq_norm_output1} or \eqref{eq_norm_output2},
\begin{equation}
	\begin{aligned}
		\mathbf{y}_L& :=\widehat{g}_L(\mathbf{x})=		
		                s_{O}^{-1}\left(
		                g_L(\gamma
		                \left( s_{I}(\mathbf{x})\right)
		                )
		                \right) \\
		\mathbf{y}_H&:= \widehat{g}_H(\mathbf{x})=
		                s_{O}^{-1}\left(
		                g_H(\gamma\left( s_{I}(\mathbf{x})\right)
		                )
		                \right) \\
	\end{aligned}	        
\end{equation} 
where $s_{O}^{-1}$ is the inverse function of the output normalization function $s_{O}$.

All in all, in our multi-fidelity architecture,  the user-defined hyperparameters are $\pmb{\theta}_U=\{L_M, \{n_i\}_{i=1}^{L_M},L_f,m,\{\sigma_i\}_{i=1}^{n_p}, d_f\}$, where $\{n_i\}_{i=1}^{L_M}$ are the number of neurons for the $L_M$ layers. 
The trainable parameters are $\pmb{\theta}=\{\pmb{\theta}_e,\pmb{\theta}_d, \pmb{\lambda}\}$, where $\pmb{\theta}_e$ and $\pmb{\theta}_d$ are the trainable parameters of the encoder net and decoder net, respectively.

\section{Training method}
\label{sec_MFTrain}
\subsection{Loss function}
In general, we are interested at phenomena that are described by:
\begin{equation}
	\begin{aligned}
		\mathcal{N}(\phi(\mathbf{x})) &= 0, \qquad  \mathbf{x} \in \Omega \\
		\mathcal{B}(\phi(\mathbf{x})) &= 0, \qquad  \mathbf{x} \in \partial \Omega \\
	\end{aligned}
\end{equation}
where $\mathcal{N}$ is a general partial differential operator defined on the domain $ \Omega$ and $\mathcal{B}$ is a general boundary condition operator defined on the boundary $ \partial \Omega$. Also, $\mathbf{y}=\phi(\mathbf{x})$ is the solution (field) which satisfies the PDE and boundary conditions. For time-dependent problems, time $t$ is considered as a  component of $\mathbf{x}$, $\Omega$ is a space-time domain, and the initial condition will be deemed as a special boundary condition of the space-time domain.

The LF output $\mathbf{y}_L$ and HF output $\mathbf{y}_H$ of the MF network are employed to approximate the LF and HF solutions of the system, respectively. The loss function of the MF network training consists of the low- and high-fidelity parts defined by the prior low- and high-fidelity  knowledge
\begin{equation}	\label{eq_lossMF}
	\mathcal{L}_{MF}(\pmb{\theta}) 
	= \mathcal{L}_{H}(\pmb{\theta}) + \mathcal{L}_L(\pmb{\theta}). 	
\end{equation}
The ``prior knowledge" could be physics, labeled data, or their combination, which is dependent on the specific problem to be solved. The low- or high-fidelity loss function is defined in a general form:
\begin{equation}
	\label{eq_lossSF}
	\begin{aligned}
		 \mathcal{L}_{\bullet}(\pmb{\theta}) 
		&= \underbrace{
			\frac{1}{N_{\bullet R}}\sum_{i=1}^{N_{\bullet R}}{\left\|\mathcal{N}(\mathbf{y}_{\bullet}(\mathbf{x}_{\bullet R}^i))\right\|^2}
			+\frac{1}{N_{\bullet B}}\sum_{i=1}^{N_{\bullet B}}{\left\|\mathcal{B}(\mathbf{y}_{\bullet}(\mathbf{x}_{\bullet B}^i))\right\|^2}
		}_{\text{Physics}}\\
		&+
		\underbrace{
			\frac{1}{N_{\bullet D}}\sum_{i=1}^{N_{\bullet D}}{\left\|\mathbf{y}_{\bullet}(\mathbf{x}_{\bullet D}^i)-\mathbf{y}_{\bullet D}^{i,*}\right\|^2}
		}_{\text{Labeled data}}  \\
	\end{aligned}
	,
\end{equation} 
where the subscript ``$\bullet$" can be ``$L$" or ``$H$" for LF and HF, respectively. $N_{\bullet R}$, $N_{\bullet B}$ and $N_{\bullet D}$ denotes the sizes of the residual data set $\mathcal{D}_{\bullet R}=\{\mathbf{x}_{\bullet R}^i\}_{i=1}^{N_{\bullet R}}$, the boundary data set $\mathcal{D}_{\bullet B}=\{\mathbf{x}_{\bullet B}^i\}_{i=1}^{N_{\bullet B}}$ and the labeled data set $\mathcal{D}_{\bullet D}=\{(\mathbf{x}_{\bullet D}^i, \mathbf{y}_{\bullet D}^{i,*})\}_{i=1}^{N_{\bullet D}}$. Note that the  low- and high-fidelity loss functions are calculated with the low- and high-fidelity outputs, respectively.

For example, for a forward problem where the prior knowledge can include high-fidelity physics and low-fidelity labeled data,  the loss function is defined as 
\begin{equation}
	\label{eq_lossMF_forward}
	\begin{aligned}
	\mathcal{L}_{MF}(\pmb{\theta}) 
	&= \mathcal{L}_{H}(\pmb{\theta}) + \mathcal{L}_L(\pmb{\theta}) \\
	&= \underbrace{
	   \frac{1}{N_{HR}}\sum_{i=1}^{N_{HR}}{\left\|\mathcal{N}(\mathbf{y}_H(\mathbf{x}_{HR}^i))\right\|^2}
	   +\frac{1}{N_{HB}}\sum_{i=1}^{N_{HB}}{\left\|\mathcal{B}(\mathbf{y}_H(\mathbf{x}_{HB}^i))\right\|^2}
	   }_{\text{High-fidelity physics}}\\
       &+
       \underbrace{
	   \frac{1}{N_{LD}}\sum_{i=1}^{N_{LD}}{\left\|\mathbf{y}_L(\mathbf{x}_{LD}^i)-\mathbf{y}_L^{i,*}\right\|^2}
	   }_{\text{Low-fidelity labeled data}}  \\
	\end{aligned}
	,
\end{equation} 
Also as an example, for an inverse problem where the prior knowledge  can include high-fidelity physics with limited number of unknown parameters $\mathcal{P}$, low- and high- fidelity labeled data, the loss function is defined as 
\begin{equation}
	\label{eq_lossMF_inverse}
	\begin{aligned}
		\mathcal{L}_{MF}(\pmb{\theta}, \mathcal{P}) 
		&= \mathcal{L}_{H}(\pmb{\theta}, \mathcal{P}) + \mathcal{L}_L(\pmb{\theta}) \\
		&= \underbrace{
			\frac{1}{N_{HR}}\sum_{i=1}^{N_{HR}}{\left\|\mathcal{N}(\mathbf{y}_H(\mathbf{x}_{HR}^i); \mathcal{P})\right\|^2}
			+\frac{1}{N_{HB}}\sum_{i=1}^{N_{HB}}{\left\|\mathcal{B}(\mathbf{y}_H(\mathbf{x}_{HB}^i); \mathcal{P})\right\|^2}
		}_{\text{High-fidelity physics}}\\
		&+
		\underbrace{
			\frac{1}{N_{HD}}\sum_{i=1}^{N_{HD}}{\left\|\mathbf{y}_H(\mathbf{x}_{HD}^i)-\mathbf{y}_{HD}^{i,*}\right\|^2}
		}_{\text{High-fidelity labeled data}}  \\	
		&+
		\underbrace{
			\frac{1}{N_{LD}}\sum_{i=1}^{N_{LD}}{\left\|\mathbf{y}_L(\mathbf{x}_{LD}^i)-\mathbf{y}_{LD}^{i,*}\right\|^2}
		}_{\text{Low-fidelity labeled data}}  \\
	\end{aligned}
	.
\end{equation} 
We note that both for forward and inverse problems, other combinations of LF and HF knowledge are possible and can be accommodated by our MF framework.

\subsection{Training with self-adaptive weighting}
\label{subsec_training}
The PINN approach proposed by \cite{raissi2019physics}, referred to herein as the baseline PINN, has witnessed a great success for predicting reasonably smooth solutions with  simple boundary conditions. However, for problems with complex features, the various terms in the loss function need to be appropriately weighted otherwise the baseline PINN can exhibit convergence and accuracy issues. Self-adaptive weighting methods \cite{wight2020solving,wang2021understanding,wang2022and,liu2021dual, mcclenny2020self} are widely used to address the issues. In the current work, we apply a recently proposed self-adaptive weighting method \cite{mcclenny2020self} for the training of the MF network. 

Taking the training of the forward problem with loss function defined in Eq. \eqref{eq_lossMF_forward} as an example, the implementation process is introduced as follows. By assigning a weight to each individual training point, the loss function in Eq. \eqref{eq_lossMF_forward} is recast as 
\begin{equation}
	\label{eq_lossMF_SA}
	\begin{aligned}
	\mathcal{L}_{MF}(\pmb{\theta}, \mathbf{w}) 
&= {
	\frac{1}{N_{HR}}\sum_{i=1}^{N_{HR}}{M(w_{HR}^i)\left\|\mathcal{N}(\mathbf{y}_H(\mathbf{x}_{HR}^i))\right\|^2}
	+\frac{1}{N_{HB}}\sum_{i=1}^{N_{HB}}{M(w_{HB}^i)\left\|\mathcal{B}(\mathbf{y}_H(\mathbf{x}_{HB}^i))\right\|^2}
}\\
&+
 {
	\frac{1}{N_{LD}}\sum_{i=1}^{N_{LD}}{M(w_{LD}^i)\left\|\mathbf{y}_L(\mathbf{x}_{LD}^i)-\mathbf{y}_{LD}^{i,*}\right\|^2}
}  \\
   \end{aligned}	
\end{equation}
where $\mathbf{w}=\{w_{HR}^i\}_{i=1}^{N_{HR}} \cup \{w_{HB}^i\}_{i=1}^{N_{HB}} \cup \{w_{LD}^i\}_{i=1}^{N_{LD}} $ are the collection of the trainable weights and $M$ is a mask function (non-negative, differentiable, increasing monotonically). In this paper, we choose the mask function $M(x)=x^2$, which is also used in \cite{mcclenny2020self}. The loss function in Eq. \eqref{eq_lossMF_SA} is minimized with respect to the parameters $\pmb{\theta}$, but maximized with respect to the weights $\mathbf{w}$, namely
\begin{equation}
	\min_{\pmb{\theta}} \max_{\mathbf{w}}  \mathcal{L}_{MF}(\pmb{\theta}, \mathbf{w}).
\end{equation}
Following a gradient ascent/descent approach, the parameters and weights are updated concurrently, namely
\begin{equation}
	\begin{aligned}
	\pmb{\theta}^{k+1}  &= \pmb{\theta}^{k} - \eta^k \nabla_{\pmb{\theta}}\mathcal{L}_{MF}^k(\pmb{\theta}, \mathbf{w}) \\
	\mathbf{w}^{k+1} &= \mathbf{w}^{k} + \rho^k \nabla_{\mathbf{w}}\mathcal{L}_{MF}^k(\pmb{\theta}, \mathbf{w}) \\
	\end{aligned} 
\end{equation}
where $k$ is the iteration number, while $\eta^k$ and $\rho^k$ are the learning rates for the parameters and self-adaptive weights, respectively. The gradient with respect to the self-adaptive weights is
\begin{equation}
	\begin{aligned} \nabla_{\mathbf{w}}\mathcal{L}_{MF}(\pmb{\theta}, \mathbf{w})
    &=
    \left\{
    \frac{1}{N_{HR}}{M^{\prime}(w_{HR}^i)\left\|\mathcal{N}(\mathbf{y}_H(\mathbf{x}_{HR}^i))\right\|^2} 
    \right\}_{i=1}^{N_{HR}}             \\
    &\cup
    \left\{
	\frac{1}{N_{HB}}{M^{\prime}(w_{HB}^i)\left\|\mathcal{B}(\mathbf{y}_H(\mathbf{x}_{HB}^i))\right\|^2} 
	\right\}_{i=1}^{N_{HB}}             \\
    &\cup
	\left\{
	\frac{1}{N_{LD}}{M^{\prime}(w_{LD}^i)\left\|\mathbf{y}_L(\mathbf{x}_{LD}^i)-\mathbf{y}_{LD}^{i,*}\right\|^2} 	\right\}_{i=1}^{N_{LD}},            	
	\end{aligned} 	\label{eq_gradien_SA}
\end{equation}
where $M'$ stands for the derivative of the mask function $M$ with respect to its argument.
Note that the above gradients can be calculated directly without the automatic differentiation to save computational cost. According to our numerical experiments, we find that the prediction accuracy of the MF neural network will be better if we remove $\frac{1}{N_R}$, $\frac{1}{N_B}$ and $\frac{1}{N_D}$ in Eq. \eqref{eq_gradien_SA}. The new gradient formulation  is as follows:
\begin{equation}
	\begin{aligned}
		\nabla_{\mathbf{w}}\mathcal{L}_{MF}(\pmb{\theta}, \mathbf{w})
		&=
		\left\{
		{M^{\prime}(w_{HR}^i)\left\|\mathcal{N}(\mathbf{y}_H(\mathbf{x}_{HR}^i))\right\|^2} 
		\right\}_{i=1}^{N_{HR}}             \\
		&\cup
		\left\{
		{M^{\prime}(w_{HB}^i)\left\|\mathcal{B}(\mathbf{y}_H(\mathbf{x}_{HB}^i))\right\|^2} 
		\right\}_{i=1}^{N_{HB}}             \\
		&\cup
		\left\{
		{M^{\prime}(w_{LD}^i)\left\|\mathbf{y}_L(\mathbf{x}_{LD}^i)-\mathbf{y}_{LD}^{i,*}\right\|^2} 
		\right\}_{i=1}^{N_{LD}} .        
	\end{aligned} 	\label{eq_gradien_SA2}
\end{equation}

The part of the training setup that is common to all the numerical experiments is as follows. The network weights of each MF network are initialized with the Xavier initialization \cite{glorot2010understanding}.  The feature shift is initialized with a Gaussian sampling $\mathcal{N}(0,0.2^2)$. The network biases are initialized as 0. The self-adaptive weights are initialized as 1.
Each network is first trained  by the Adam optimizer \cite{kingma2014adam} for 72000 iterations, followed by the L-BFGS  optimizer \cite{liu1989limited} for 8000 iterations. It is noted that the training with the L-BFGS optimizer  is necessary and can  improve the accuracy significantly \cite{raissi2019physics}. For the Adam training, the learning rate $\eta$ for the parameters  is initialized as 0.001, and decreases by 1\% every 400 iterations.  The self-adaptive weights are only updated during the Adam part of the training with a fixed learning rate $\rho=0.1.$ They are kept constant during the subsequent L-BFGS part of the training.

\section{Results and discussion}
\label{sec_R_D}
To test the performance of the proposed multi-fidelity network architecture,  three kinds of problems are tested, namely steady-state  problems, time-dependent problems and inverse problems. According to  Section \ref{subsec_MF_summary}, except the regular hyperparameters of both encoder and decoder networks, there still exist some key hyperparameters in the MF architecture, namely the average wave numbers of Fourier feature embedding, the feature depth and the feature distance. First, the influence of these hyperparameters will be studied with the steady lid-driven flow problem, based on the control variates method, where only the investigated variable is varied while the other variables are kept unchanged with baseline. Then, the performance of the MF architecture will be further studied for time-dependent problems and inverse problems.

In order to show the improved performance of the MF approach, the results of the following two PINN approaches are also presented for comparison:
\begin{enumerate}[(1)]
	\item \textbf{Single HF}:  The MF network is trained with only high-fidelity  knowledge, where the number of residual and boundary points as well as their distribution laws are the same with those of the MF approach. 
	If applicable, the locations of the high-fidelity data points are the same with those used for the MF approach. Its loss function is only the   high-fidelity part in Eq. \eqref{eq_lossMF}, namely  $\mathcal{L}_{H}$.	
	\item \textbf{HF with data}:  This approach is only applied for forward problems to provide an accuracy limit for the MF approach by replacing the LF labeled data with the high-fidelity data. In this approach, the MF network is trained with both physics and HF data, where the number of residual and boundary points as well as their distribution laws are the same with those in the MF approach, the locations of the high-fidelity data points are the same with those of LF data points in the MF approach. Its loss function is the sum of the physics part in Eq. \eqref{eq_lossMF_forward} and a HF data part $\mathcal{L}_{HD}(\pmb{\theta})$:
	\begin{equation}
		\mathcal{L}_{HD}(\pmb{\theta}) = 
		\frac{1}{N_{LD}}\sum_{i=1}^{N_{LD}}{\left\|\mathbf{y}_H(\mathbf{x}_{LD}^i)-\mathbf{y}_{HD}^{i,*}\right\|^2},
	\end{equation}
	with the exact data set $\mathcal{D}_{HD}=\{(\mathbf{x}_{LD}^i, \mathbf{y}_{HD}^{i,*})\}_{i=1}^{N_{LD}}$. 
\end{enumerate}
Note that the MF network is also employed for the HF approach and the HF with data approach, for the purpose of a fair comparison. Since the two approaches are of single fidelity, only the HF output of the MF network are used in their loss functions. Besides, the networks of the two approaches are also trained with the self-adaptive weighting method described in Section \ref{subsec_training}.

In the MF approach, we train the networks with physics and additional labeled data  of low fidelity. Physics can provide a more accurate knowledge to the HF output, while the low-fidelity data can restrict the HF output to a small neighborhood around itself. Thus the MF prediction is expected to be better than the networks trained with physics only. In the case of forward problems, the  HF with data  approach becomes prohibitively expensive  for real-world applications, due to the cost of generating abundant HF data. Besides, training with only the exact data could be faster and more accurate than the HF with data approach. However, the performance of the HF with data approach can be thought of as an ideal reference for the MF approach.  Due to the use of exact data, the prediction accuracy of the HF with data approach is expected to be higher than the MF approach, and thus the accuracy of the  HF with data  approach can be deemed as the maximum accuracy limit of the MF approach. 

To measure the prediction accuracy, the relative $L2$  error is employed:
\begin{equation}
	\varepsilon_{\bullet}
	=\sqrt{
		\frac
		{\sum\limits_{\mathbf{x},\mathbf{y}\in \mathcal{D}_T}
			{\left\| \mathbf{y}_{\bullet}(\mathbf{x})-\mathbf{y} \right\|_2^2}
		}
		{\sum \limits_{\mathbf{x},\mathbf{y}\in \mathcal{D}_T}
			{\left\| \mathbf{y} \right\|_2^2}
		}
	}
	\label{eq_relative_error}
\end{equation}
where $ \mathcal{D}_T =\{\left(\mathbf{x}_T^i,\mathbf{y}_T^i\right)\}_{i=1}^{N_T}$ is the test data set of size $N_T$, the subscript ``$\bullet$" can be ``LF", ``Single HF", ``MF" or ``HF with data".

The corresponding training of neural networks are all implemented in PyTorch \cite{paszke2019pytorch}, and run on a server-class supercomputer using 32-bit single-precision data type and single GPU (NVIDIA $^\circledR$ Corporation TU102).

\subsection{Steady lid-driven cavity flow problem}
{\label{subsec_lid_steady}}
In this subsection, the lid-driven cavity flow problem is used to test the performance of the proposed MF approach and also the influence of several hyperparameters. The  flow  enclosed in a square cavity $\Omega=[0,1]^2$ is described by the (non-dimensional) incompressible Navier-Stokes equations
\begin{equation}
\left \{
\begin{aligned}
\nabla \cdot {\mathbf{u}} &= 0,  &\mathbf{x} &\in \Omega\\
 {\mathbf{u}} \cdot \nabla {\mathbf{u}} &=  - \nabla p + \frac{1}{Re}{\nabla^2}{\mathbf{u}}, & \mathbf{x} &\in \Omega \\
\mathbf{u}(\mathbf{x})&=(u_w(\mathbf{x}), 0),  & \mathbf{x} &\in \Gamma _1  \\
 \mathbf{u}(\mathbf{x})&=0,     & \mathbf{x} & \in \Gamma _2 \\
\end{aligned}
\right .
,
\label{eq_GoverningEqsLidDriven}
\end{equation}
where $\mathbf{u}=(u,v)$ is the velocity in the Cartesian coordinate system $\mathbf{x}=(x,y)$, $p$ is the pressure and $Re$ is the Reynolds number. The boundary is $\partial \Omega=\Gamma _1 \cup \Gamma _2$, where $\Gamma _1$ represents the top moving lid and $\Gamma _2$ represents the other three static non-slip walls.
$u_w$ is the driving velocity of the moving lid. To overcome the  singularity at the two upper corner points where the moving lid meets the two stationary vertical walls, a zero-corner-velocity profile $u_w$ is employed \cite{phillips1993treatment, chen2018multidomain}:
\begin{equation}
u_w(\mathbf{x}) = 16x^2(1-x)^2
\end{equation}

To obtain the training data for networks, the low- and high-fidelity numerical simulations are conducted with  a Chebyshev pseudo-spectral numerical solver \cite{chen2018multidomain}. 
The high-fidelity solution (considered as exact solution) is simulated with a fine mesh resolution of $61 \times 61$, while the low-fidelity solutions are simulated with a  relatively low resolution. The resolution adopted is given in Table \ref{table_params_steday_lid}. The low- and high-fidelity data sets are collected from the points on a $51\times51$  uniform grid with their corresponding values interpolated (using Chebyshev spectral expansion) from the low- and high-fidelity solutions, respectively. The residual data set is made up of $N_R$ residual points randomly sampled in the domain $\Omega$. The test data set is collected from the points and their values in the high-fidelity solution. The data set sizes are given in Table \ref{table_params_steday_lid}.

\begin{table}[tbp]
	\newcommand{\tabincell}[2]{\begin{tabular}{@{}#1@{}}#2\end{tabular}}
	\centering
	\caption{Resolution in simulations and data set sizes for steady lid-driven flow problem.}
	\begin{tabular}{ccccccc}
		\toprule
		\multirow{2}{*}{ $Re$}  &  \multicolumn{2}{c}{Resolution $(Nx,Ny)$} & \multirow{2}{*}{ $N_{HR}$} & \multirow{2}{*}{ $N_{HB}$}  & \multirow{2}{*}{ $N_{LD}$}  & \multirow{2}{*}{ $N_{T}$}\\
		\cmidrule(lr){2-3}
		&      LF & Exact     \\
		\midrule
		400 & (11,11)    & \multirow{4}{*}{(61,61)}   & 4096 &\multirow{4}{*}{400} &\multirow{4}{*}{2601}
		&\multirow{4}{*}{3721}\\
		1000 & (15,15)   & &4096  &&& \\
		2500 & (21,21)   & &8192  &&&\\
		5000 & (27,27)   & &16384 &&&\\
		\bottomrule
	\end{tabular}
	\label{table_params_steday_lid}
\end{table}

The baseline user-defined hyperparameters for building the multi-fidelity network are as follows. The encoder net and decoder net are built with 6 layers, each layer containing 50 neurons. The feature depth is $L_f=6$, implying the 6th layer is chosen as the feature layer. The feature distance is chosen as $d_f=1.0$. The Fourier feature embedding for the input is constructed with
$m=100$ Fourier features and  the average wave number $\pmb{\sigma}=\{\sigma_x, \sigma_y\}=\{0.5,0.5\}$ for $x$ and $y$ directions.

\begin{figure}[htbp]
	\centering	
	\subfigure[]{
		\includegraphics[width=7.2cm]{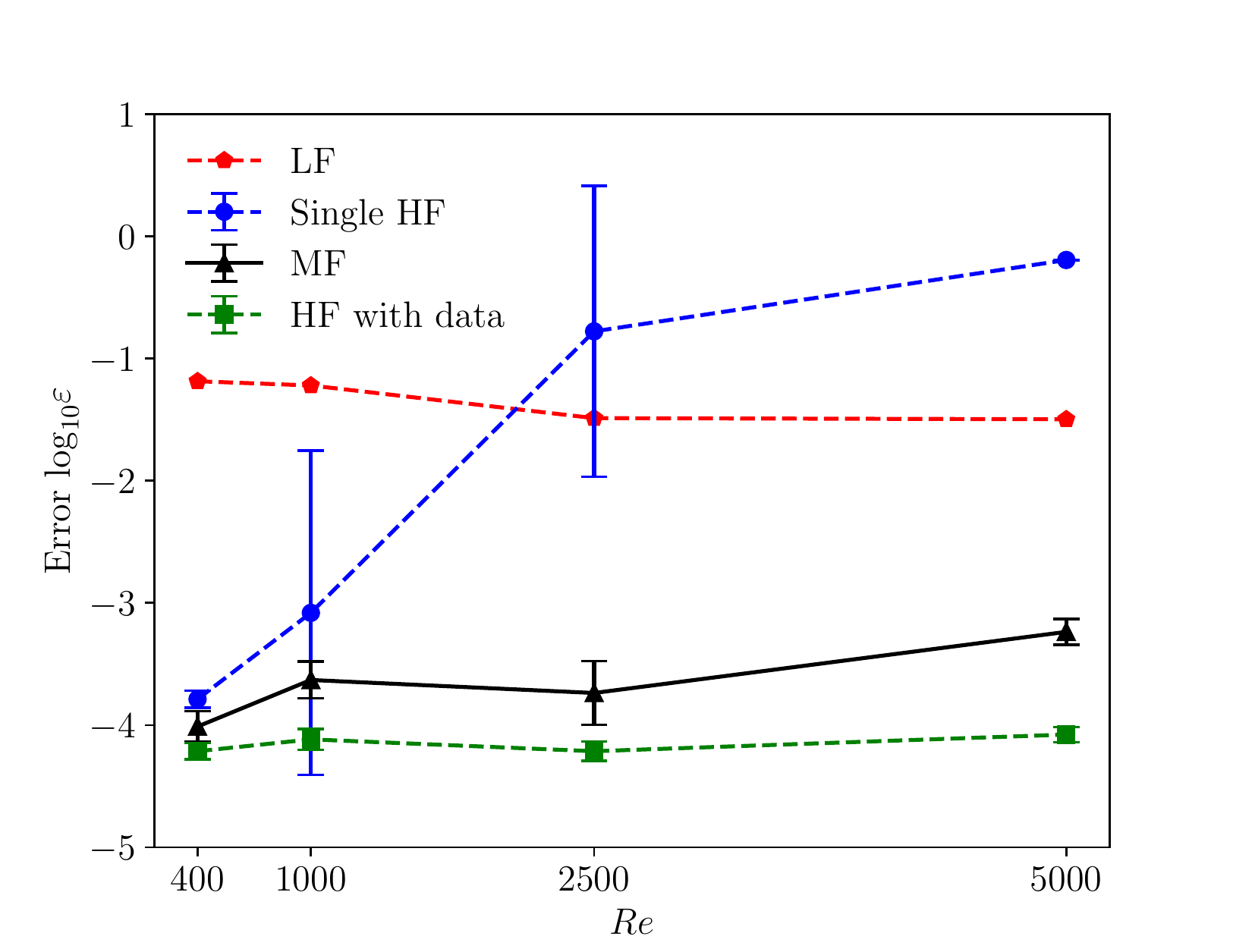}
	}	
	\subfigure[]{
		\includegraphics[width=7.2cm]{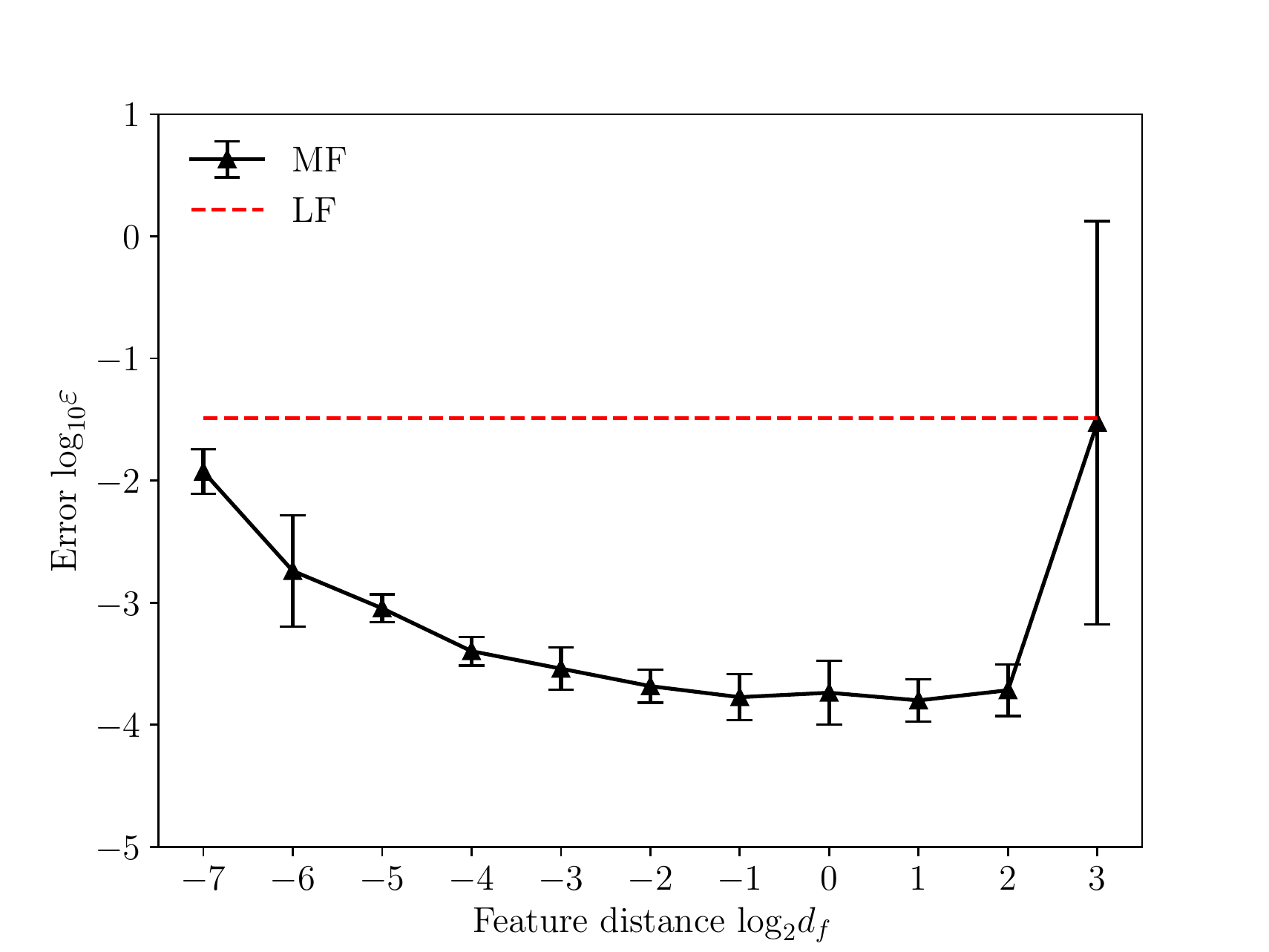}
	}
	
	\subfigure[]{
		\includegraphics[width=7.2cm]{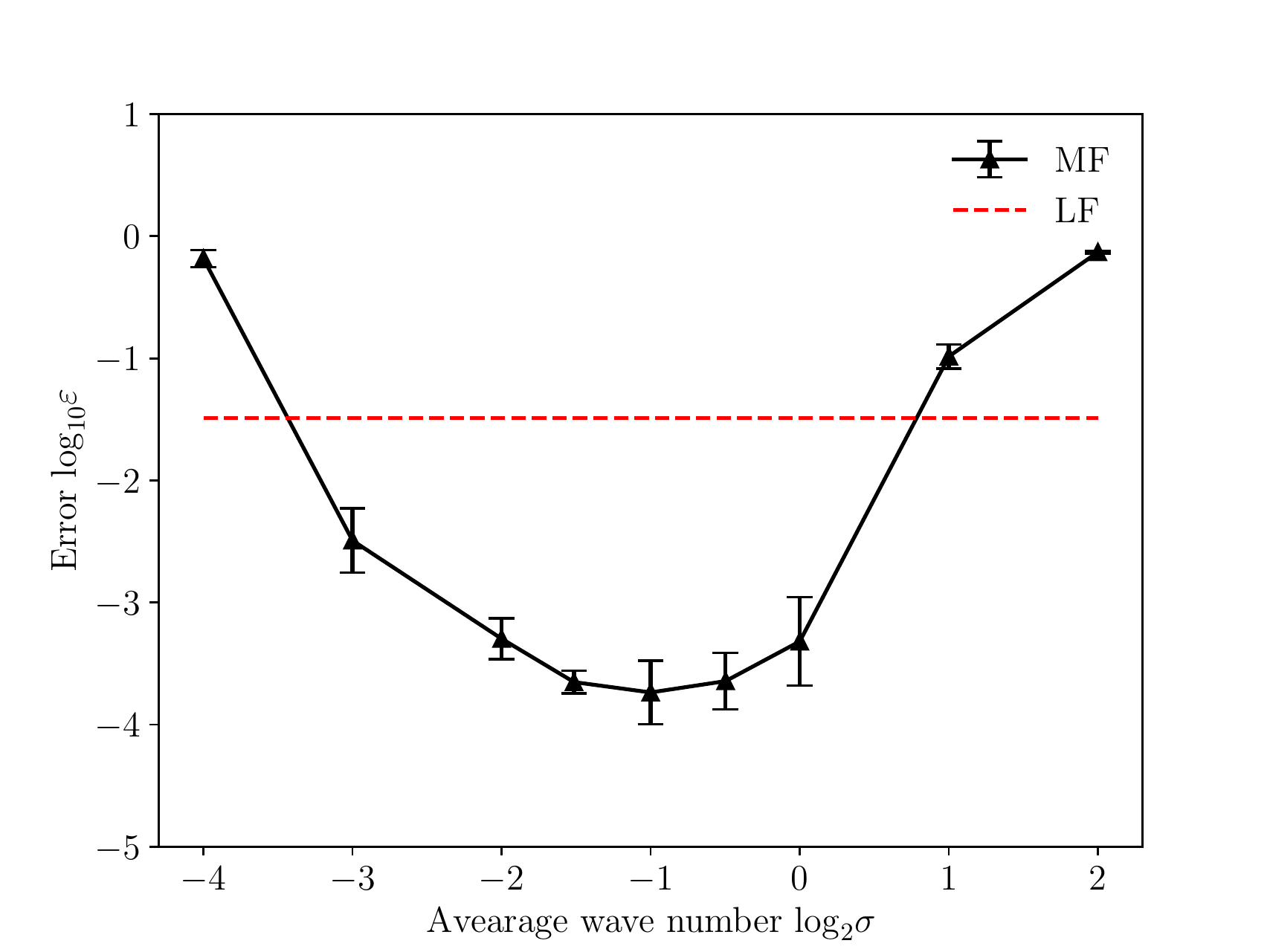}
	}
	\subfigure[]{
		\includegraphics[width=7.2cm]{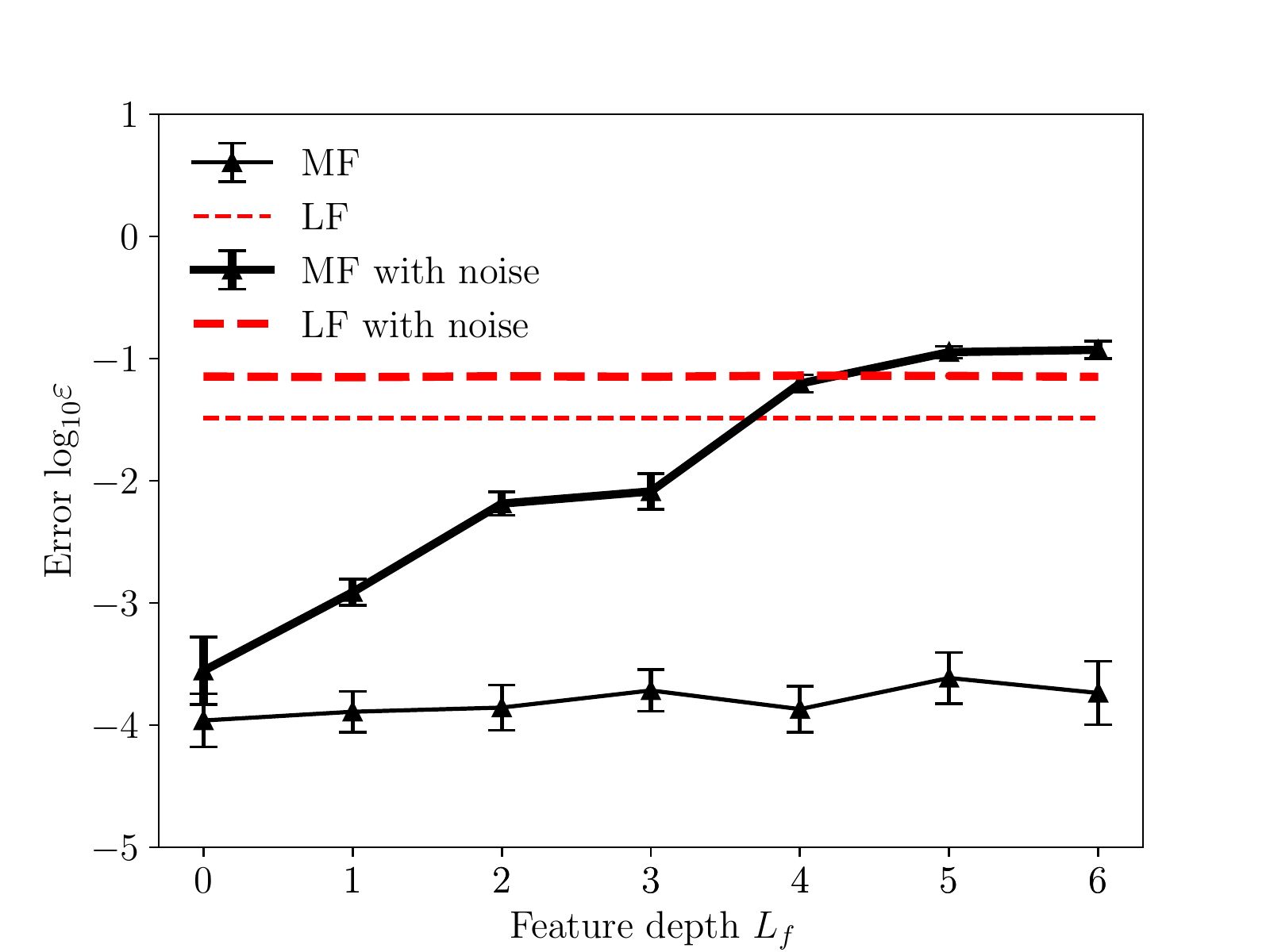}
	}
	\caption{Steady lid-driven flow problem: the prediction accuracy of networks with respect to (a) Reynolds number $Re$, (b) the feature distance $d_f$, (c) the average wave number $\sigma_x=\sigma_y=\sigma$ in Fourier feature embedding and (d) the feature depth $L_f$. The mean and standard deviation are calculated over 5 independent runs. }
	\label{fig_liddriven_accuray}
\end{figure}

\begin{figure}[H]
\centering	
\includegraphics[width=14.4cm]{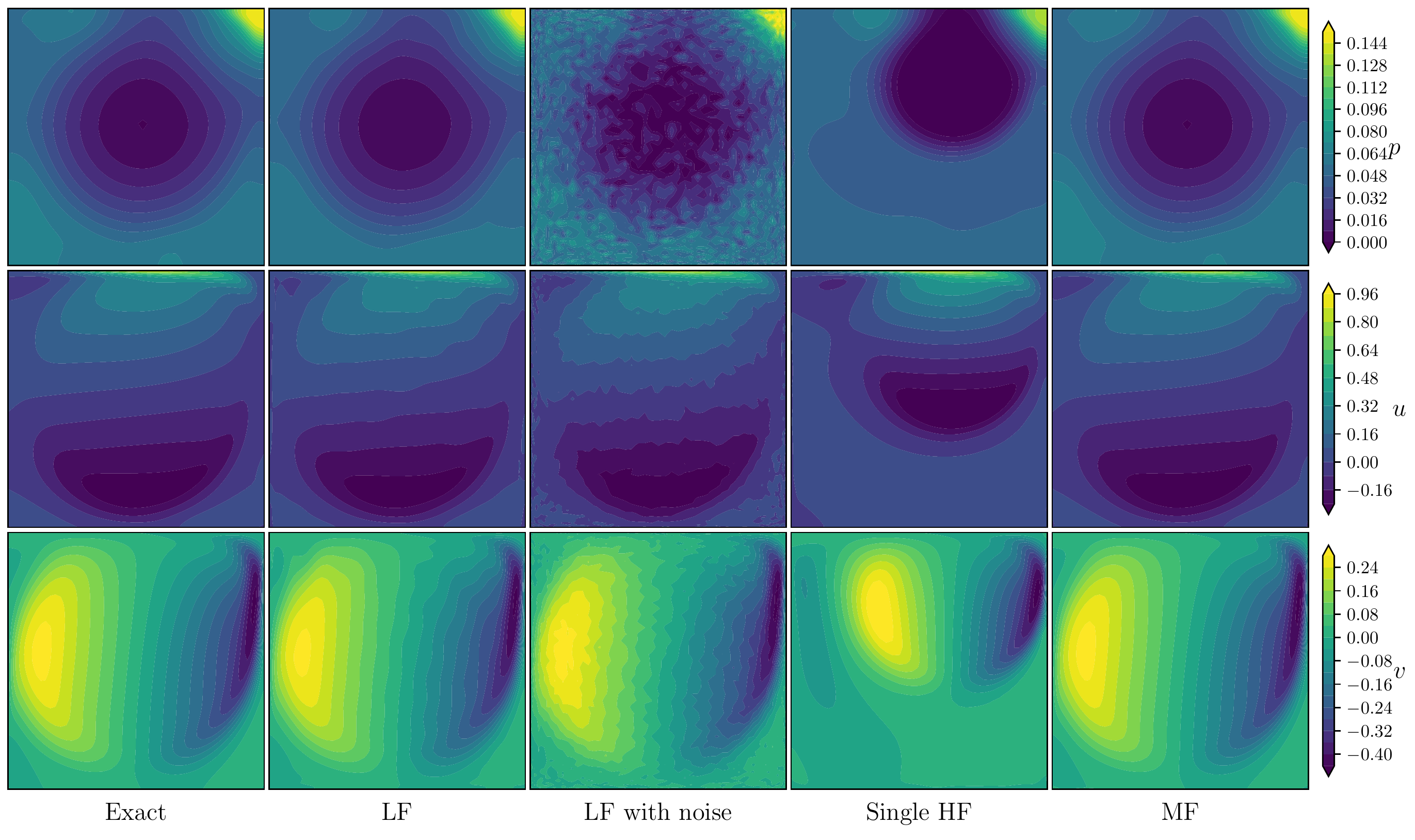}
\caption{Steady lid-driven flow: the predictions of  the Single HF and the MF approaches versus the reference high-fidelity (Exact) solution, the noisy low-fidelity (LF with noise) solution and the low-fidelity  (LF) solution for $Re=2500$.}
\label{fig_liddriven_contourfRe2500}	
\end{figure}

First, the prediction accuracy of the MF network is studied for $Re$=400, 1000, 2500, 5000. Note that $Re=5000$ is close to the critical Reynolds number (about 7500 \cite{bruneau20062d}) where the first Hopf bifurcation occurs.  As shown in Fig. \ref{fig_liddriven_accuray},  the single HF approach is not accurate for $Re > 1000$, its relative error becomes close to 100\% for larger $Re$ values. However, with the guidance of the low-fidelity solutions, the relative error of the MF approach is $(1.01 \pm 0.30)\times10^{-4}, ( 2.50 \pm 0.95)\times10^{-4}, (2.24 \pm 1.60)\times10^{-4}, (5.96\pm 1.27)\times10^{-4}$ for $Re=400, 1000, 2500, 5000$, respectively,
which is about two orders of magnitude lower than the low-fidelity solution. Also, the relative error of the MF approach is only a little higher than its minimum limit, namely the error of the HF with data approach. 
The prediction results of the MF and single HF approaches for $Re=2500$ are shown in Fig. \ref{fig_liddriven_contourfRe2500}. The single HF approach fails to predict the boundary layer along the right sidewall. The boundary layer thickness is overestimated, implying that the single HF approach struggles to predict the large gradients within the boundary layers. 
But for the MF approach, the low-fidelity data can  provide an approximated boundary layer profile, guiding the physics part to approximate the exact solution.

\renewcommand{\dblfloatpagefraction}{.9}
 %\vspace{-0.5cm}

Second, the influence of the user-defined  feature distance,  feature depth and  average wave number in the Fourier feature embedding  on the MF prediction accuracy  is investigated. The Reynolds number is fixed at $Re=2500$. When studying the influence of a hyperparameter, all the other hyperparameters are fixed as those in the baseline user-defined hyperparameters. The results are shown in Fig. \ref{fig_liddriven_accuray}. It is seen that the MF prediction accuracy is not sensitive to the feature distance, implying a wide feasible range for the feature distance. But the feature distance should not be too large, otherwise, the low-fidelity solution will lose effect in guiding the high-fidelity physics. Of course, it also should not be too small, otherwise the MF prediction will be attracted to the low-fidelity solution. As for the average wave number in  Fourier feature embedding, its  accuracy curve has a similar trend with that in \cite{tancik2020fourier}. The average wave number should also not be too small/large. Otherwise, the prediction will be either under-fitting or over-fitting, as pointed out in \cite{tancik2020fourier}. The average wave number should be chosen carefully, due to its relatively narrow feasible range. According to our empirical tests, it is recommend that $\sigma_i \in [1/3 K_i, 1/2 K_i]$  promises a better prediction, where $K_i$ denotes the  wave number of the main flow. As for the feature depth, it is interesting that it has minimal influence on the MF prediction accuracy. This implies that the network training process is able to build an encoder/decoder net to match both low-fidelity and high-fidelity solutions. 

We also consider one kind of noisy low-fidelity data set, which is generated by adding a white noise $\Delta \sim\mathcal{N}(0,0.01)$ to the high-fidelity solution, as shown in Fig.  \ref{fig_liddriven_contourfRe2500}. With the noisy low-fidelity data set, the prediction accuracy of the MF approach decreases with the increase of the feature depth. Since white noise contains high-frequency modes, we suspect that the feature-adjacent space derived from smaller feature depth has richer frequencies, and thus the feature shift acts as a filter to remove/weaken noisy high-frequency modes.

\renewcommand{\dblfloatpagefraction}{.9}
% \vspace{-0.5cm}

\subsection{Time-dependent problems}
In time-dependent problems, one issue for the baseline PINN is its difficulty in addressing short timescales \cite{wang2021understanding}. Increasing the time span of a simulation is a  way to make time evolution faster with respect to the normalized time. Therefore, in this subsection, we will mainly consider the influence of time span on prediction accuracy of the multi-fidelity approach.
\subsubsection{Pendulum problem}
\label{subsec_Pendulem}
As our first example of time-dependent problems we use a gravity pendulum with damping defined by the following two-dimensional ordinary differential equation (ODE) system
\begin{equation}
	\left\{
	\begin{aligned}
		\frac{\partial s_1}{\partial t} &= s_2, \qquad &t&\in [0,T]\\
		\frac{\partial s_2}{\partial t} &= -\frac{b}{m}s_2-\frac{g}{L}\sin(s_1), \qquad &t&\in [0,T]\\	
		s_1(0)&=s_2(0)=1 \\
	\end{aligned}
	\right .
	,
\end{equation}
where $T$ is the time span to be solved. In this example, we take $m=L=1,b=0.05$ and $g=9.81$, as tested in \cite{wang2023long}.

To obtain the training data for the neural networks, the low- and high-fidelity simulations are conducted by using the  standard four-stage Runge-Kutta method for explicitly marching the solution in time. The time steps employed in the low- and high-fidelity simulations are $\Delta t=1/3$ and $\Delta t=1/100$, respectively. The low- and high-fidelity data sets are collected from the time points in the low-fidelity simulation and values taken from the low- and high-fidelity solutions, respectively. The residual data set are collected  from $2^{14}T/100$ uniform time points. The test set are collected from the time points with their values from the high-fidelity simulation.

The user-defined hyperparameters for building the multi-fidelity network are as follows. The encoder net and decoder net are built with 6 layers, each layer containing 50 neurons. The feature depth is $L_f=6$, implying the 6th layer is chosen as the feature layer. The feature distance is chosen as $d_f=1.0$. The Fourier feature embedding for the input is constructed with $m=100$ Fourier features and  the average wave number ${\sigma}=5T/100$.

\begin{figure}[htbp]	
	\centering	
	\includegraphics[width=8cm]{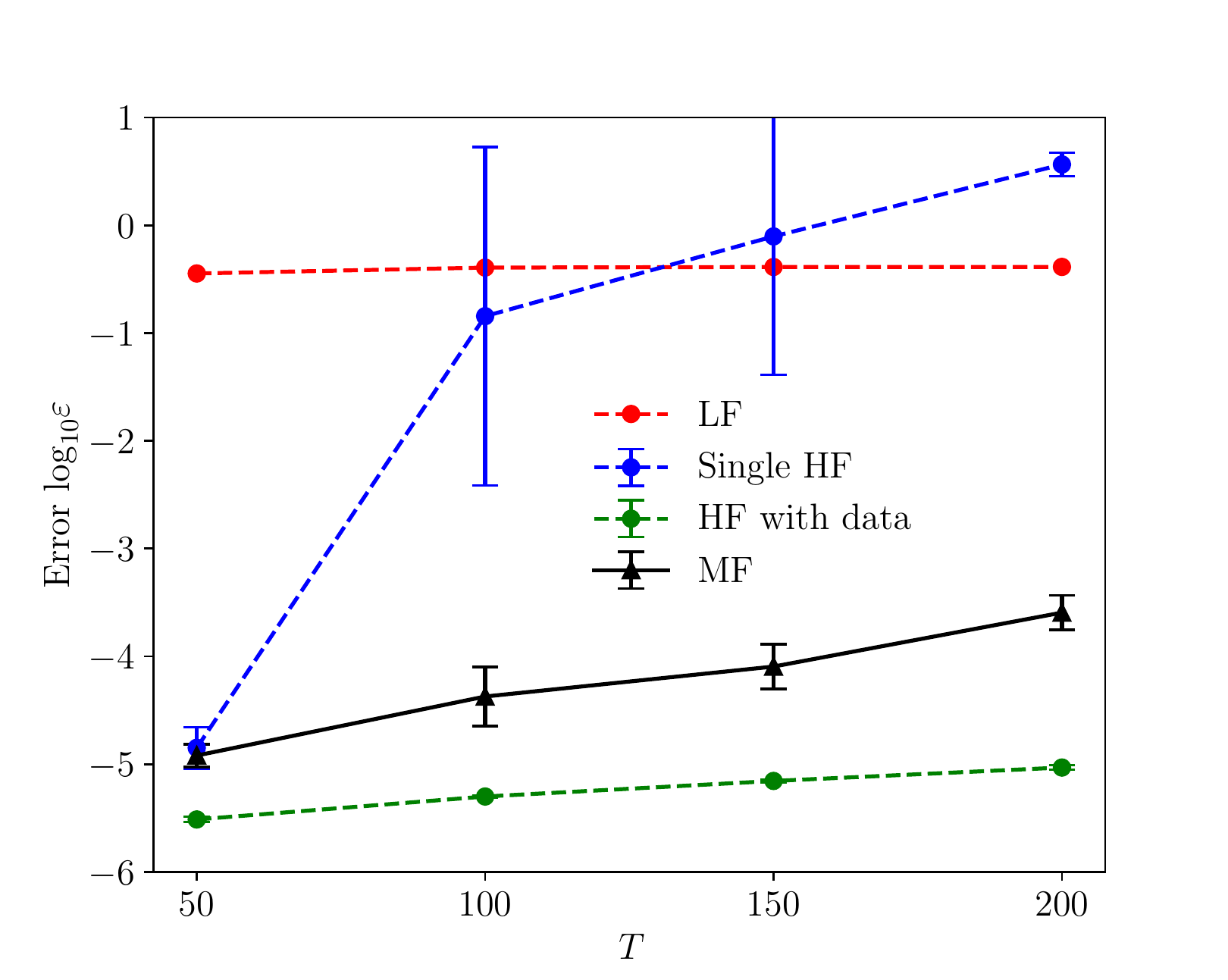}
	\caption{Pendulum problem: the relative error of the various approaches versus that of the reference  high-fidelity (Exact) solution and low-fidelity  solution. The mean and standard deviation are calculated over 5 independent runs.}
	\label{fig_pendulum_accuracy}	
\end{figure}

\begin{figure}[htbp]		
	\centering	
	\includegraphics[width=14.4cm]{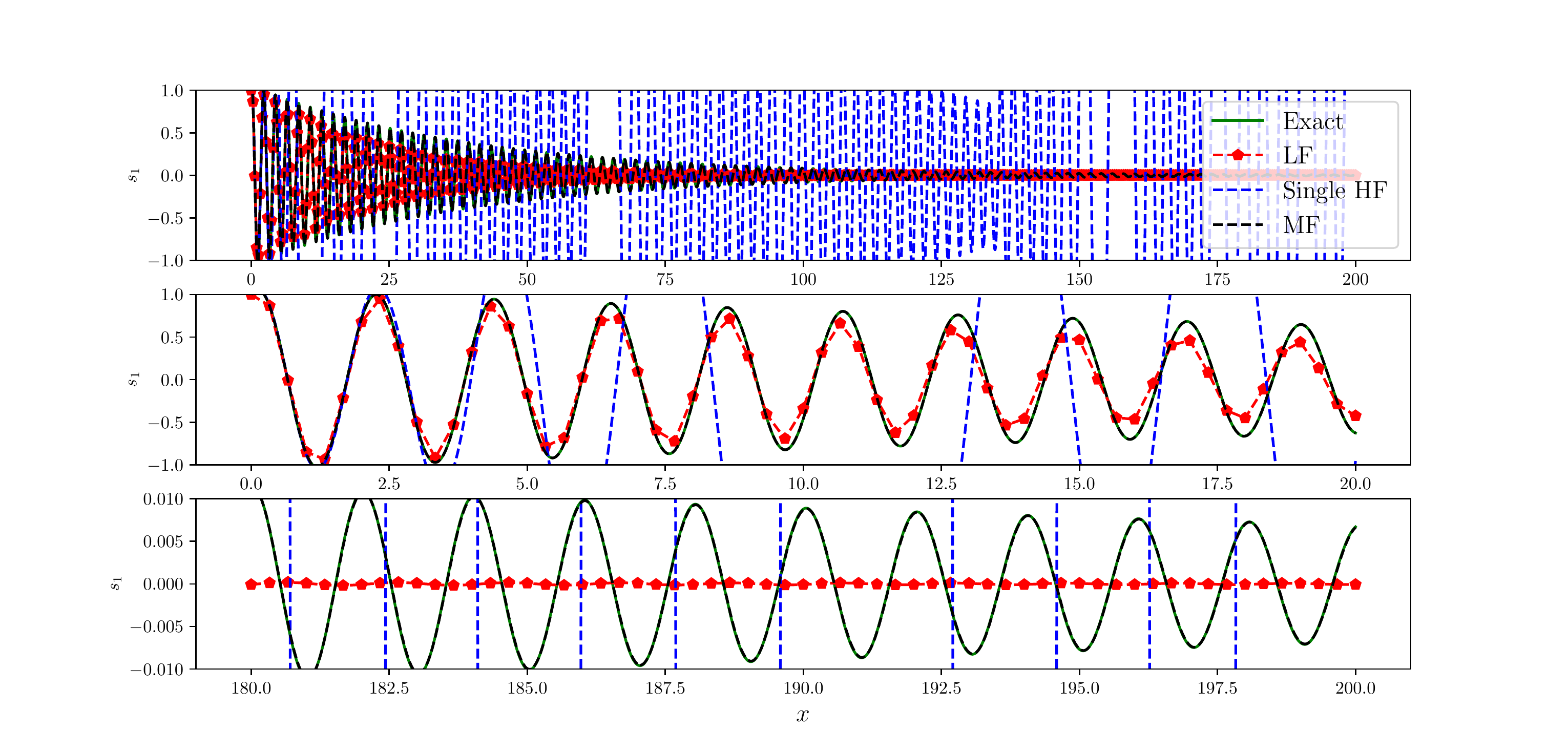}
	\caption{Pendulum problem: the predictions  of the MF approach and the single HF approach versus the reference  high-fidelity (Exact) solution and low-fidelity  solution.}
	\label{fig_pendulum_history}
\end{figure}

The relative error (prediction accuracy) for the time span $T=50,100,150,200$ is shown in Fig. \ref{fig_pendulum_accuracy}. The accuracy of the single HF approach reduces substantially with the increase of the time span. However, the MF approach offers a much more accurate prediction, with just a slight accuracy degradation for the increasing time span.
The relative error of the MF approach is $(1.24 \pm 0.35) \times 10^{-5}$, $(5.28 \pm 3.86) \times 10^{-5}$, $(9.11 \pm 4.11) \times 10^{-5}$, $(2.71 \pm 0.91) \times 10^{-4}$ for $T=50, 100, 150, 200$, which is
 about 3 orders of magnitude lower than that of the low-fidelity solution.
The prediction of the solution state $s_1$ for $T=200$ is shown in Fig. \ref{fig_pendulum_history}. The low-fidelity solution $s_1$  collapses to 0 after $T=100$. However, the low-fidelity solution can still guide the MF approach to find the right solution. A possible explanation is that the low-fidelity solution can inform the high-fidelity physics of the oscillation frequency and the mean/amplitude values of each solution state. On the contrary, the single HF approach fails in predicting the right solution and develops a non-physical and irregular oscillation.

 %\vspace{-0.5cm}

 %{-0.5cm}

\subsubsection{Gray-Scott problem}
\label{subsec_GS}
The Gray-Scott problem is studied to test further the performance of the MF approach for the case of nonlinear time-dependent PDE problems. This is an advection-diffusion-reaction problem, governed by the following one-dimensional nonlinear PDE system
\begin{equation}
	\label{eq_GS_gov1}
	\left\{
	\begin{aligned}
	\frac{\partial u}{\partial t} &= 
	r_u \frac{\partial^2 u}{\partial x^2}
	-uv^2+f(1-u), &(x,t)&\in[-L,L]\times[0,T]\\
	\frac{\partial v}{\partial t} &= 
	r_v \frac{\partial^2 v}{\partial x^2}
	+uv^2-(f+k)v, &(x,t)&\in[-L,L]\times[0,T] \\
	u(x,0)&=1-\frac{1}{2} \sin^2 (\pi \frac{x-L}{2L}), &x&\in[-L,L] \\
	v(x,0)&=	\frac{1}{4} \sin^2 (\pi \frac{x-L}{2L}),  &x&\in[-L,L] \\	
	u(x,t)&=u(x+2L,t), &(x,t)&\in \mathbb{R} \times [0,T] \\	
	\end{aligned}
	\right .
	,
\end{equation}
where $T$ is the time span to be solved. In this example, we take $r_u=1, r_v=0.1, f=0.1, k=0$ and $ L=50$. Since the solution is symmetric about $x=0$ and periodic along $x$ axis, we only solve the right half part $[0,L]\times[0,T]$ to reduce the computational cost. The periodic boundary condition in Eq. \eqref{eq_GS_gov1} is replaced with 
\begin{equation}
	\label{eq_GS_gov2}
	\frac{\partial u(0,t)}{\partial x}=\frac{\partial u(L,t)}{\partial x}=0, \qquad t \in[0,T] 
	.
\end{equation}

To obtain the training data for neural networks, the low- and high-fidelity simulations are performed using the  Chebfun library \cite{driscoll2014chebfun} with  the Fourier spectral method for spatial discretization and a fourth-order stiff time-stepping scheme for marching in time. 
The number of Fourier modes employed in the low- and high-fidelity simulations are 16 and 512, respectively. The time steps employed in the low- and high-fidelity simulations are $\Delta t=2$ and $\Delta t=1$, respectively. The low- and high-fidelity data sets are collected from the space-time points in the low-fidelity simulation with their values from the low- and high-fidelity simulations, respectively. The residual data set is generated by randomly sampling $200T$ points in the space-time domain. The test data set is collected from the space-time points and the corresponding solution values in the high-fidelity simulation.

The user-defined hyperparameters for building the multi-fidelity network are as follows. The encoder net and decoder net are built with 6 layers, each layer containing 50 neurons. The feature depth is $L_f=6$, implying the 6th layer is chosen as the feature layer. The feature distance is chosen as $d_f=1.0$. The Fourier feature embedding for the input is constructed with $m=100$ Fourier features and  the average wave numbers $\pmb{\sigma}=\{0.5,0.5T/100\}$ for $x$ and $t$ directions.

\begin{figure}[htbp]
	\centering	
	\includegraphics[width=8cm]{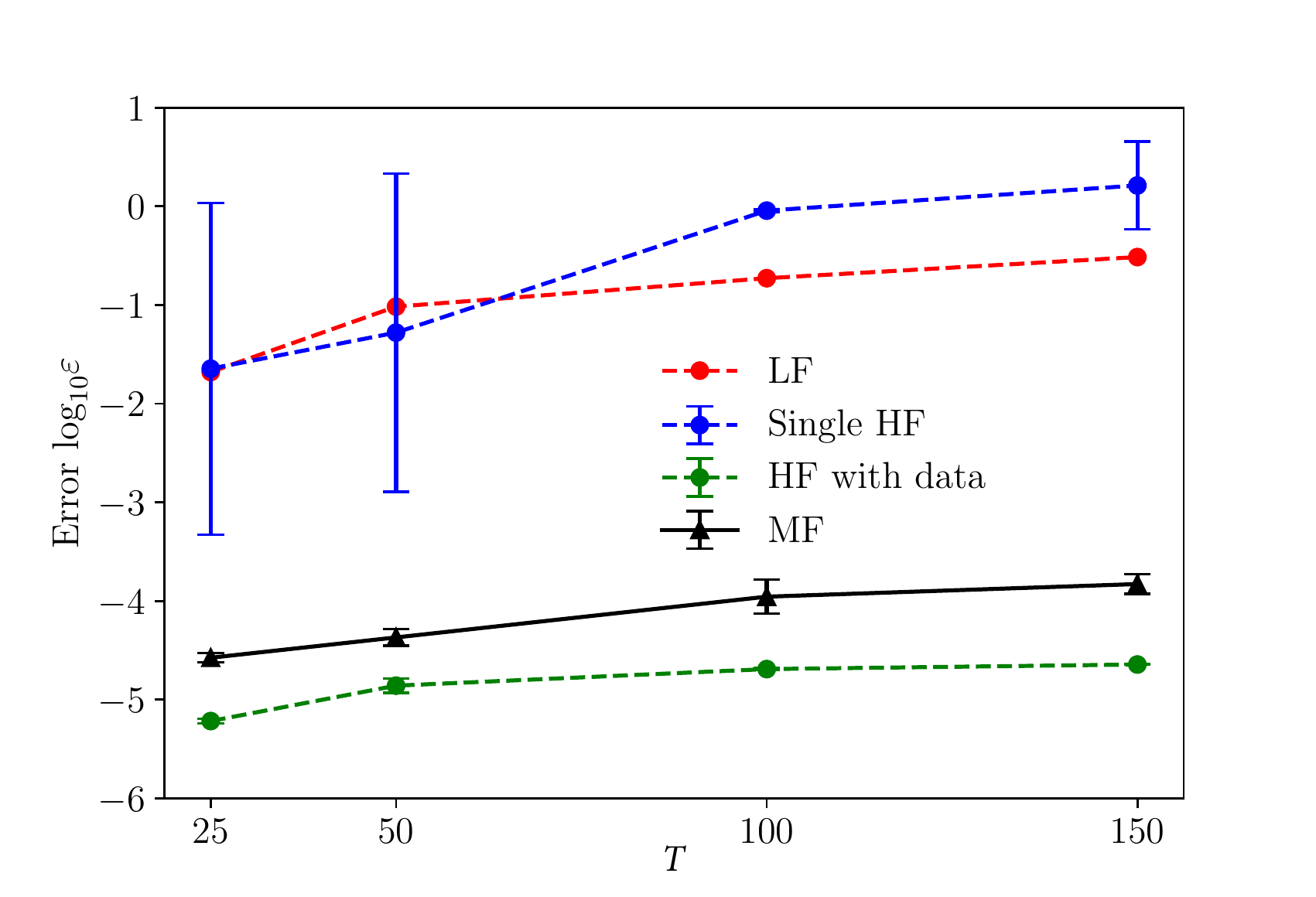}
	\caption{Gray-Scott problem: the prediction accuracy of networks versus that of the reference  high-fidelity (Exact) solution and low-fidelity  solution. The mean and standard deviation are calculated over 5 independent runs.}
	\label{fig_GrayScott_accuracy}		
\end{figure}

\begin{figure}[htbp]
	\centering	
	\subfigure{
	\includegraphics[width=14.4cm]{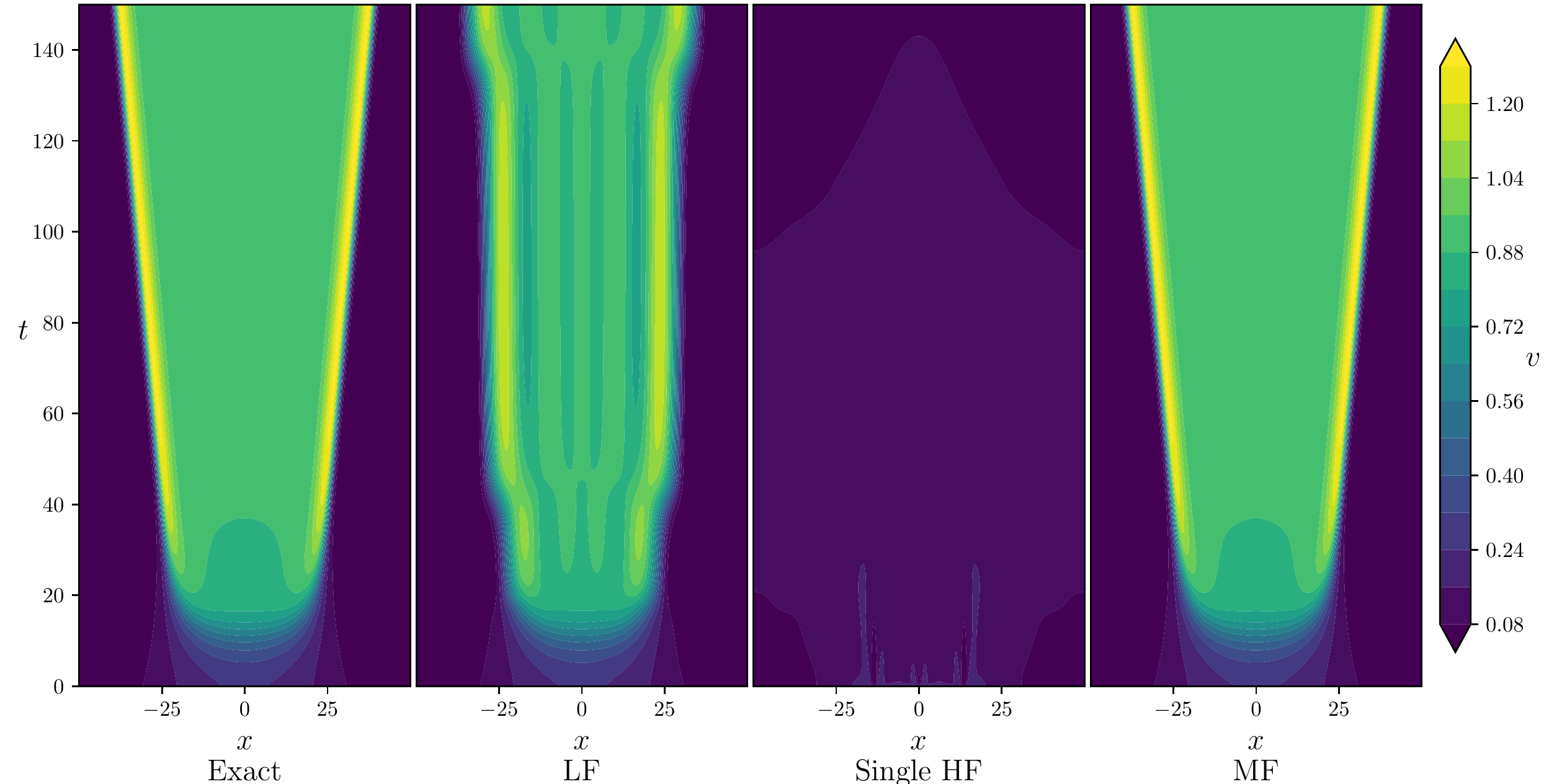}
	}	
	\subfigure{
	\includegraphics[width=14.4cm]{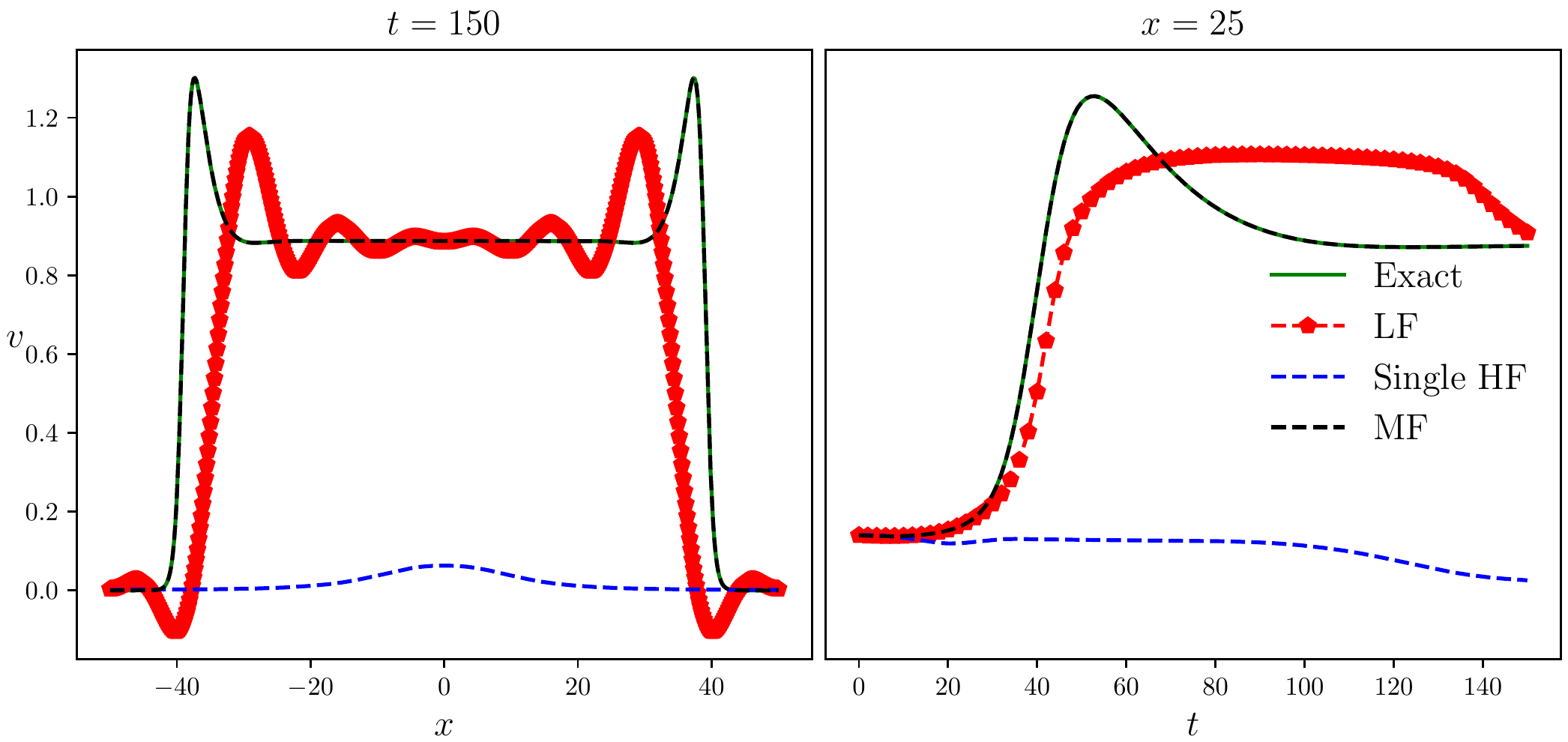}
	}	
	\caption{Gray-Scott problem: (top) global view and (bottom) section view of  the predictions  of the MF approach and the single HF approach versus the reference  high-fidelity (Exact) solution and the low-fidelity  solution.}
	\label{fig_GrayScott_contourf}		
\end{figure}

The relative error (prediction accuracy) for the time span values $T=25,50,100,150$ is shown in Fig. \ref{fig_GrayScott_accuracy}. The accuracy of the single HF approach degrades significantly with the increase of the time span.
Even for the shortest time space $T=25$, the single HF approach is unstable with a large variance for prediction accuracy. 
On the contrary, the MF approach provides a much more accurate prediction with small variance, and only a slight accuracy degradation with the increase of the time span. For $T=25,50,100,150$, the relative error of the MF approach is $(2.68 \pm 0.29)\times10^{-5},( 4.36 \pm 0.91)\times10^{-5}, (1.19 \pm 0.43)\times10^{-4}, (1.53\pm 0.36)\times10^{-4}$, respectively.
The prediction of the solution $v$ is shown in Fig. \ref{fig_GrayScott_contourf}. The low-fidelity solution $v$ diverges away from the high-fidelity solution after $t=20$, showing a shift of the two ridges and an oscillation around the two ridges. This is resulting from  underestimating the contribution of the high-frequency Fourier modes. 
However, the low-fidelity solution can still guide the MF approach to find the right solution. The physics part in the MF approach is able to complement the high-frequency Fourier modes.  The prediction of the single HF approach is rather poor. This is because the single HF approach is unable to resolve the two sharp ridges. A possible explanation of the success of the MF approach is that the low-fidelity solution is not accurate but contains useful information about the locations of the ridges. Since there is no such prior information for the single HF approach, it gets trapped into the problem of ``spectral bias".

 %\vspace{-0.5cm}

 %\vspace{-0.5cm}

\subsubsection{Unsteady lid-driven flow problem}
To further validate the performance of the MF approach, the unsteady lid-driven flow problem is considered. The flow is governed by the following 2D PDE system:
\begin{equation}
	\left \{
	\begin{aligned}
		\nabla \cdot {\mathbf{u}} &= 0  &(\mathbf{x},t) &\in \Omega \times [0,T]\\
		\frac{\partial \mathbf{u}}{\partial t} +
		{\mathbf{u}} \cdot \nabla {\mathbf{u}} &=  - \nabla p + \frac{1}{Re}{\nabla^2}{\mathbf{u}}  &(\mathbf{x},t) &\in \Omega \times [0,T] \\
		\mathbf{u}(\mathbf{x}, t)&=(u_w(\mathbf{x},t), 0)  & (\mathbf{x},t) &\in \Gamma _1 \times [0,T] \\
		\mathbf{u}(\mathbf{x}, t)&=0     & (\mathbf{x},t) & \in \Gamma _2 \times [0,T] \\
		\mathbf{u}(\mathbf{x}, 0) &= \mathbf{u}_0(\mathbf{x})  &\mathbf{x} &\in \Omega\\
	\end{aligned}
	\right .
	, 
	\label{eq_GoverningEqsUnsteadyLidDriven}
\end{equation}
where $T$ is the time span and Reynolds number is $Re=1000$. The geometry and boundary conditions are same with those in Section \ref{subsec_lid_steady}, except that the velocity of the moving lid is is enforced with a periodic oscillation (in time), namely
\begin{equation}
	u_w(\mathbf{x},t) = 16x^2(1-x)^2 + \sin(2\pi t)\sin(2\pi x).
\end{equation}

To obtain the training data for the neural networks, the low- and high-fidelity simulations are conducted with the Chebyshev pseudo-spectral method for spatial discretization and the second-order backward difference method for marching in time \cite{chen2020parallel}. 
The spatial resolution employed in the low- and high-fidelity simulations are (15,15) and (61,61), respectively. The initial solution $\mathbf{u}_0$ of the high-fidelity simulation is the steady-state solution at $Re=1000$ from Section \ref{subsec_lid_steady}. The initial solution of the low-fidelity simulation is interpolated (using Chebyshev spectral expansion) from that of the high-fidelity simulation.  The time steps employed in the low- and high-fidelity simulations are $\Delta t=0.05$ and $\Delta t=0.002$, respectively. To generate the low- and high-fidelity data sets, the low- and high-fidelity solutions at each time slice are first interpolated (using Chebyshev spectral expansion) to a $51\times51$  uniform grid. Then, the low- and high-fidelity data sets are collected from $10000T$ randomly chosen space-time points with solution values from the low- and high-fidelity interpolations, respectively. The residual data set is generated by randomly sampling $20000T$ space-time points in the space-time domain. The test set is collected from $10^6$ randomly chosen space-time points with their values taken from the high-fidelity simulation.

The user-defined hyperparameters for building the multi-fidelity network are as follows. The encoder net and decoder net are built with 6 layers, each layer containing 50 neurons. The feature depth is $L_f=6$, implying the 6th layer is chosen as the feature layer. The feature distance is chosen as $d_f=1.0$. The Fourier feature embedding for the input is constructed with $m=200$ Fourier features and  the average wave numbers $\pmb{\sigma}=\{0.5,0.5,0.5T\}$ for the $x$, $y$ and $t$ directions.

\begin{figure}[htbp]
	\centering	
	\subfigure[]{
		\includegraphics[width=7.2cm]{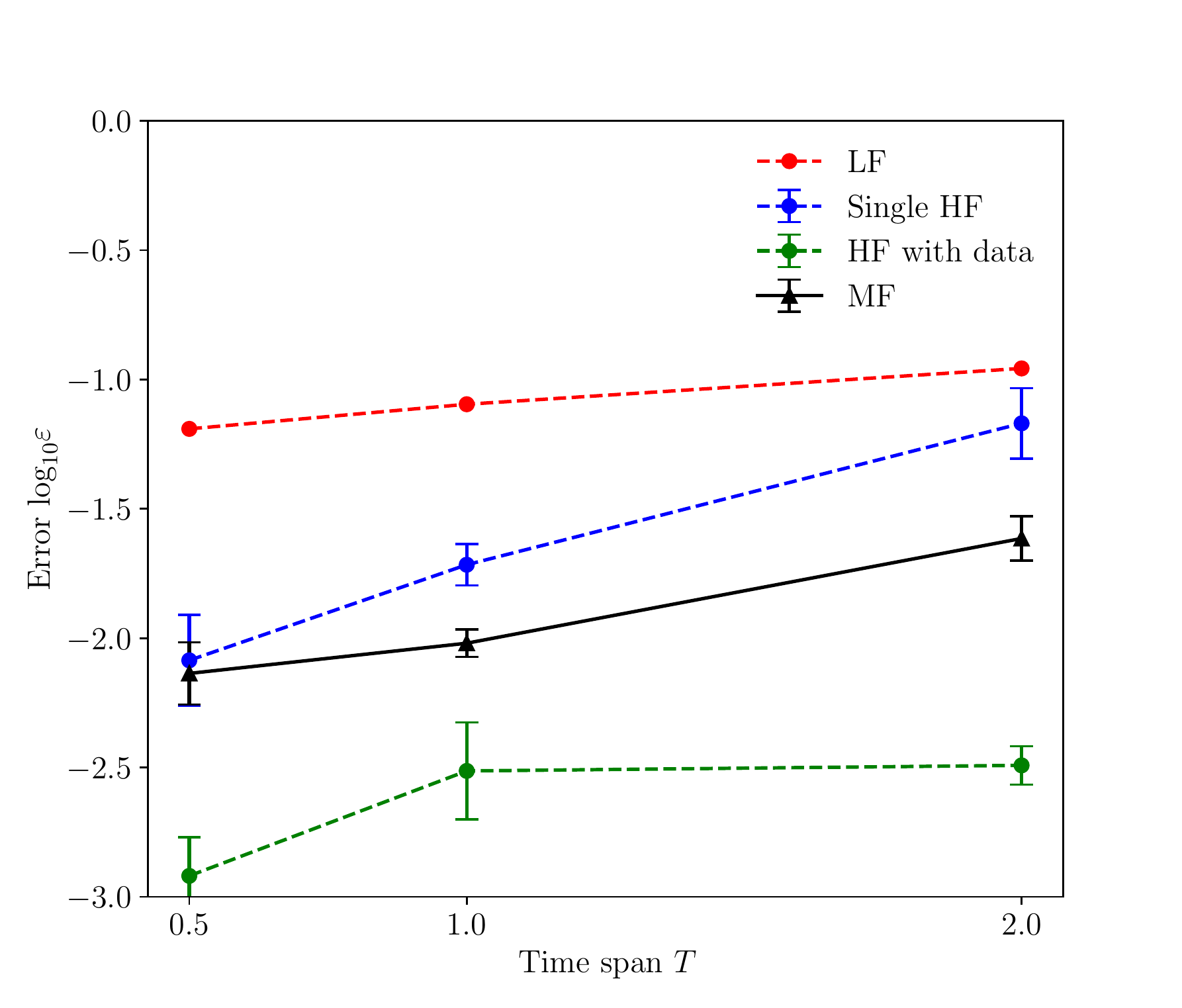}
	}	
	\subfigure[]{
		\includegraphics[width=7.2cm]{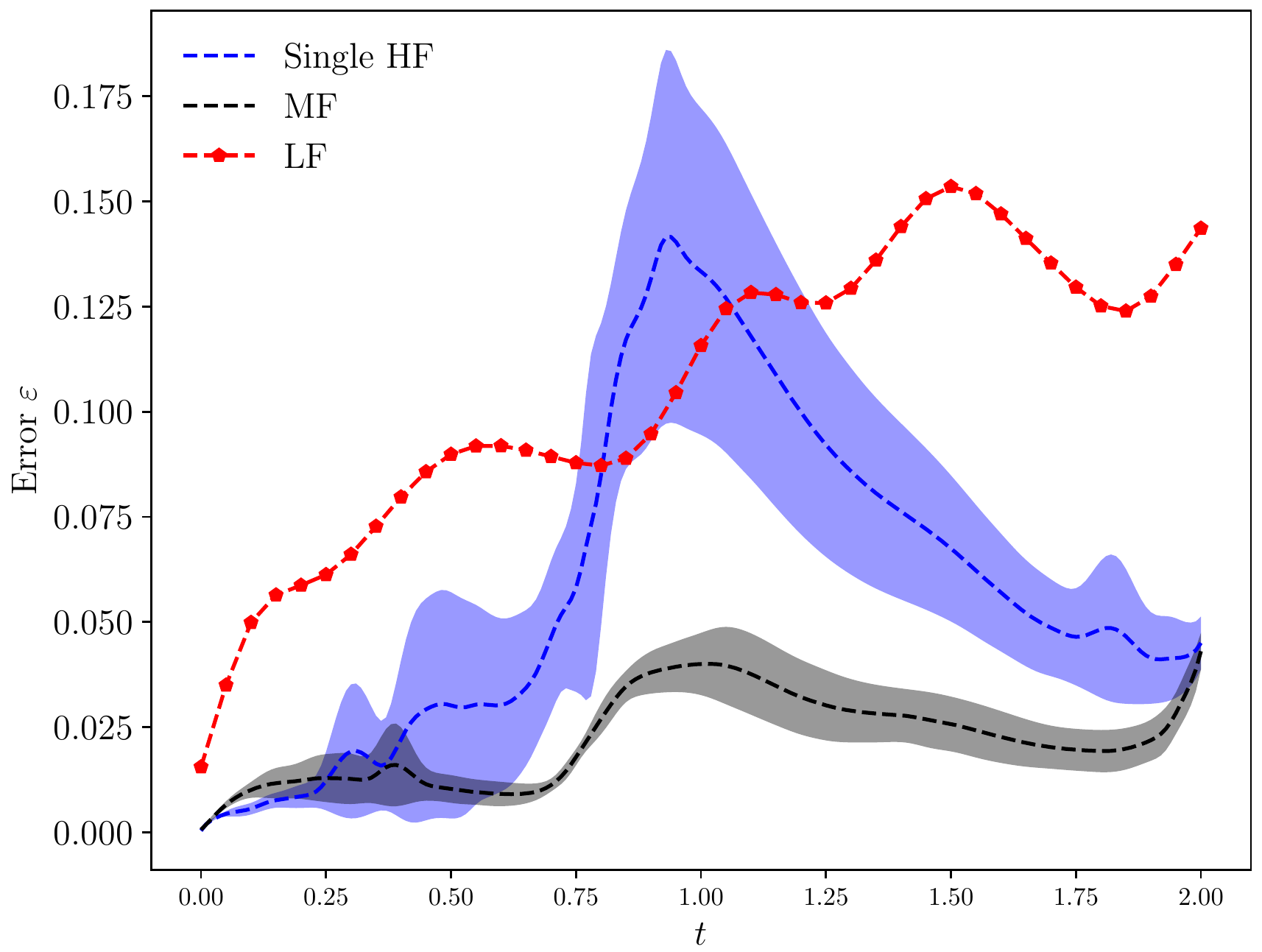}
	}
	\caption{Unsteady lid-driven flow problem: (a) global (in-time) relative error and (b) local (in-time) relative error for each time slice for $T=2.0$. The mean and standard deviation are calculated over 5 independent runs.}
	\label{fig_unsteadyLiddriven_accuray}
\end{figure}

\begin{figure}[htbp]
	\centering	
	\includegraphics[width=14.4cm]{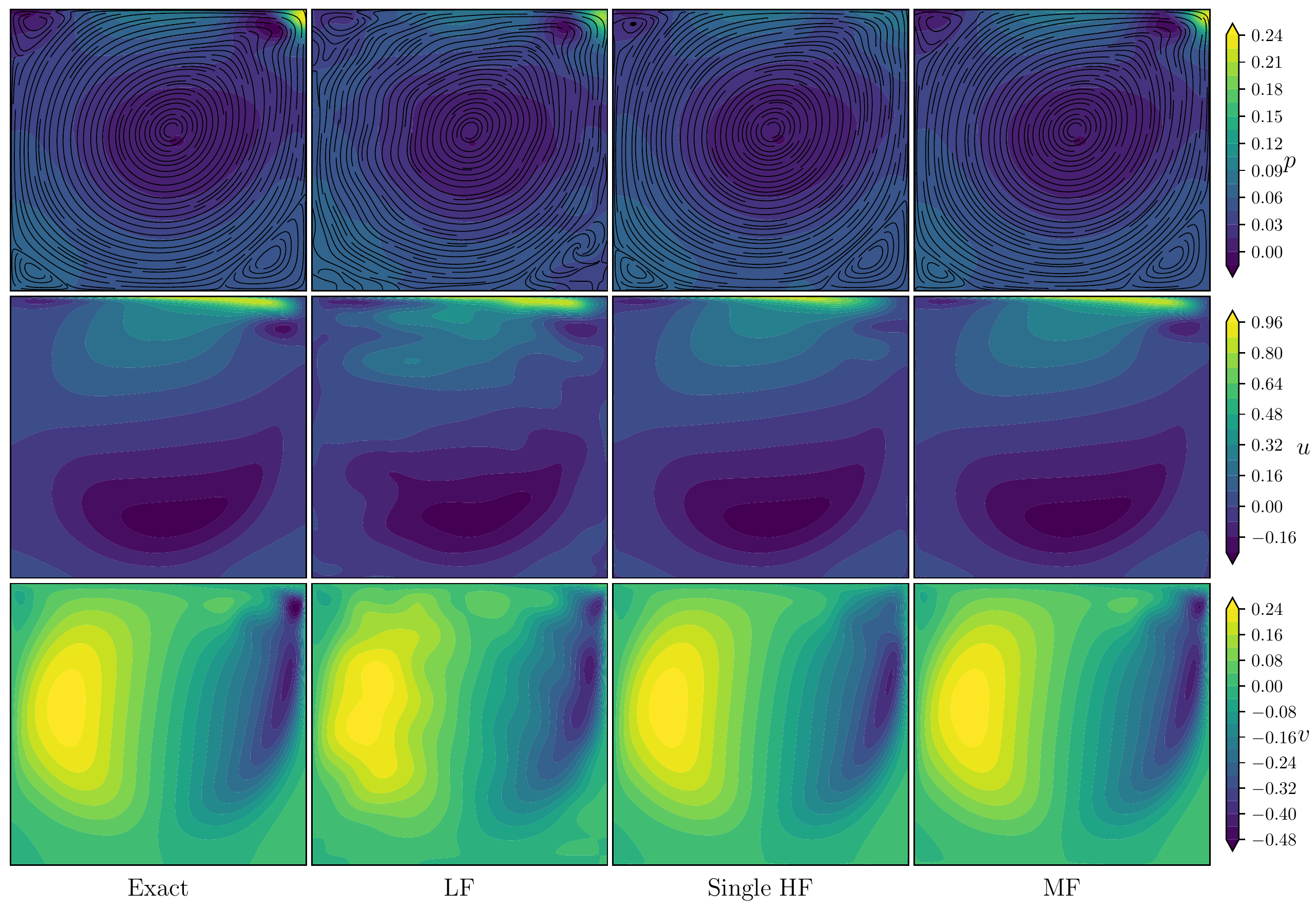}
	\caption{Unsteady lid-driven flow problem: the predictions of  the Single HF and the MF approaches versus the reference high-fidelity (Exact) solution and  low-fidelity  (LF) solution at $t=1.0$ for time span $T=2.0$. In the first row, black solid line is the streamline.}
	\label{fig_unsteadyLiddriven_contourf}
\end{figure}

The relative error (prediction accuracy) for the time span $T=0.5, 1.0,2.0$ is shown in Fig. \ref{fig_unsteadyLiddriven_accuray}(a). The relative error of the MF approach is $(7.60 \pm 2.22)\times 10^{-3}$, $(9.63 \pm 1.81)\times 10^{-3}$, $(2.47 \pm 0.47)\times 10^{-2}$ for $T=0.5, 1.0, 2.0$ respectively, which is about one order of magnitude lower than that of the low-fidelity solution.
The accuracy of the single HF approach decreases more quickly than the MF approach with the increase of time span. For the time span $T=2.0$, the relative error of the MF approach is only about half an order of magnitude lower than that of the single HF approach. The accuracy improvement of the MF approach over the single HF approach is not as significant as those in Sections \ref{subsec_Pendulem} and \ref{subsec_GS}.
However, we find that the relative error of the single HF prediction increases significantly with time for  $t<1.0$, as shown in Fig. \ref{fig_unsteadyLiddriven_accuray}(b). Taking a look at the flow details at $t=1.0$ in Fig. \ref{fig_unsteadyLiddriven_contourf}, it is seen that the single HF prediction neglects the streamline bump (characterized by a pressure minimum) near the top-right corner. On the contrary, the MF approach can  predict accurately the streamline bump, based on the guidance from the low-fidelity solution. We conclude that the MF approach outperforms the single HF approach in capturing important flow details.
 
\renewcommand{\dblfloatpagefraction}{.9}
 %\vspace{-0.5cm} 

\renewcommand{\dblfloatpagefraction}{.9}
 %\vspace{-0.5cm} 

\subsection{Inverse problems}
Although the training method introduced in Section \ref{sec_MFTrain} is intended for the forward problems, its extension to inverse problems is straightforward. For all the following inverse problems, the unknown parameters $\mathcal{P}$ to be discovered are trained in the same way as the network parameters $\pmb{\theta}$, namely they are trained first by Adam optimizer (with the same learning rate $\eta$) and then by L-BFGS optimizer, if not stated otherwise.

\subsubsection{Learning the hydraulic conductivity for nonlinear unsaturated flows}
Unsaturated flows are central in characterizing ground-surface water interaction \cite{markstrom2008gsflow,hayashi2002effects}. We consider a 1D steady unsaturated flow, which is governed by the following 1D  differential equation
\begin{equation}
\label{eq_gov_hydraulic}
        \left \{
	\begin{aligned}
	\frac{d(-K(h) \frac{dh}{dx})}{dx}&=0, \qquad &x&\in[0,L] \\
	h&=h_0, &x&=0 \\
	h&=h_1, &x&=L \\
	\end{aligned}
        \right .
,
\end{equation}
where $h$ is the pressure head, $h_0$ is the pressure head at the inlet, $h_1$ is the pressure head at the outlet. Here we take $h_0=-3cm$, $h_1=-10cm$, and $L=200cm$ as adopted in the case II in \cite{meng2020composite}. $K(h)$ denotes the pressure-dependent hydraulic conductivity, which is defined as
\begin{equation}
	\begin{aligned}
	S_e(h)&=\frac{1}{\left(1+|\alpha_0h|^{\left(\frac{1}{1-m}\right)}\right)^m} \\
K(h)&=K_sS_e^{1/2}\left[1-\left(1-S_e^{1/m}\right)^m\right]^2 \\
	\end{aligned}
,
\end{equation}
where the loam is selected as the soil type, $K_s=1.04cm/hr$ is the saturated hydraulic conductivity, $S_e$ is the effective saturation defined by the van Genuchten model \cite{van1980closed}, $\alpha_0$ is an empirical parameter related to the inverse of air entry suction, and $m$ is an empirical  parameter related to the pore-size distribution.
The parameters $\alpha_0$ and $m$ are difficult to measure because of the complex geometry of  porous media. We want to estimate the unknown empirical parameters $\alpha_0$ and $m$ based on the observations of $h$. 

The exact values of $\alpha_0$ and $m$ are assumed to be 0.036 and 0.36, respectively. To obtain the training data, numerical simulations are conducted with the built-in function $bvp5c$ in MATLAB$^\circledR$ on a uniform grid of mesh size $\Delta x=8$. The high-fidelity simulations are conducted with the exact values of $\alpha_0$ and $m$. The low-fidelity simulations are conducted with a value of  $\alpha_0$ and $m$,  uniformly sampled in the empirical ranges [0.015, 0.057] and [0.31, 0.40] for loamy soil \cite{van1980closed}, respectively. The low-fidelity observations are chosen as 26 uniformly sampled points with the solution values taken from the low-fidelity simulation. The high-fidelity observations are chosen at 2 fixed points ($x=96$ and $x=184$) from the high-fidelity simulation. $400$ uniformly sampled points are chosen as residual points. 
As considered in \cite{meng2020composite}, the inlet flux $q_0$ is also assumed to be known, thus Eq. \eqref{eq_gov_hydraulic} can be recast into the  form
\begin{equation}
        \left \{
	\begin{aligned}
		-K(h)\frac{dh}{dx}&=q_0, \qquad &x&\in[0,L] \\
		h&=h_0, &x&=0 \\
		h&=h_1, &x&=L \\
	\end{aligned}
        \right .
	.
	\label{eq_gov_hydraulic2}
\end{equation}
Note that the new governing equation is over-determined, and it is only used as a physics constraint for network training.

The user-defined hyperparameters for building the multi-fidelity network are as follows. The encoder net and decoder net are built with 4 layers, each layer containing 10 neurons. The feature depth is $L_f=4$, implying the 4th layer is chosen as the feature layer. The feature distance is chosen as $d_f=1.0$. For this case, since the solution curve is smooth,  we do not use  Fourier feature embedding. In addition, only the Adam part of  training is employed for this problem.

In network training, $\alpha_0$ and $m$ are randomly initialized from the empirical ranges [0.015, 0.057] and [0.31,0.40], respectively. Note that we infer $\alpha_0$ and $m$ indirectly by learning their scaled values with the scaling transformation
\begin{equation}
    \widetilde{\alpha_0} = \frac{\alpha_0-(0.057+0.015)/2}{(0.057-0.015)/2},
    \qquad 
    \widetilde{m} = \frac{m-(0.40+0.31)/2}{(0.40-0.31)/2}.
\end{equation}
The inferred values for $\alpha_0$ and $m$  of the present work and available works in the literature are given in Table \ref{table_accuracy_hydraulic}. The predictions of the single HF approach and the MF approach are  extremely accurate with a small relative error and a tiny standard deviation.

We also tried a more challenging scenario by directly using the differential form in Eq. \eqref{eq_gov_hydraulic} rather than the form in Eq. \eqref{eq_gov_hydraulic2}. Note that the inlet flux is unknown. We use the same network setup, but use 2000 residual points for training the networks. The inferred values from the single HF approach and the MF approach are also given in Table  \ref{table_accuracy_hydraulic}. The predictions of the MF  approach are better than those of the HF approach. The low-fidelity observations help the MF approach to  be more robust, achieving a smaller relative error.

\begin{table}[tbp]
	\newcommand{\tabincell}[2]{\begin{tabular}{@{}#1@{}}#2\end{tabular}}
	\centering
	\caption{Summary of  inferred parameters of the single HF approach and the MF approach  for hydraulic conductivity. The mean and standard deviation are calculated over 10 independent runs. * denotes the predictions are obtained by using the governing equation of differential form.}
	\begin{tabular}{ccccc}
		\toprule
		Method &  $\alpha_0$ & Error & $m$ & Error \\
		\midrule
		Exact & 0.036  & - & 0.360  & -\\
		Ref. \cite{meng2020composite} with Single HF & 0.0440  & 22.22\% & 0.377  & 4.72\% \\
		Ref. \cite{basir2022physics}  with Single HF & 0.0351 $\pm 7.18\times 10^{-4}$ & 2.58\% &0.354 $\pm 2.78\times 10^{-3}$ & 1.78\% \\
		Present with Single HF & $\mathbf{0.0360 \pm 9.80\times 10^{-7}}$ &  \textbf{0.03\%} & $\mathbf{ 0.360  \pm 3.00\times 10^{-6}}$ & \textbf{0.01\%} \\
		*Present with Single HF & 0.0359 $\pm 1.26\times 10^{-5}$ & 0.12\% & 0.360  $\pm 1.73\times 10^{-4}$ & 0.18\% \\		
		Ref. \cite{meng2020composite} with MF & $ 0.0337 \pm  7.91\times 10^{-4}$ & 6.39\% & $0.349 \pm 3.70\times 10^{-3}$ & 3.06\% \\
		Ref. \cite{basir2022physics} with MF & $0.0359 \pm 7.51\times 10^{-4}$ & 0.30\% & $0.357 \pm 2.74 \times 10^{-3}$ & 0.86\% \\
		Present with MF & $\mathbf{0.0360  \pm 6.02\times 10^{-6}}$ & \textbf{0.02\%} & $\mathbf{0.360 \pm 5.38\times 10^{-6}}$ & \textbf{0.01\%} \\
		*Present with MF & $0.0360  \pm 3.65\times 10^{-5}$ & 0.07\% & $0.360 \pm 5.33\times 10^{-4}$ & 0.11\% \\		
		\bottomrule
	\end{tabular}
	\label{table_accuracy_hydraulic}
\end{table}

\subsubsection{Parameter inference for a chemical reaction model}
We further consider a 1D chemical reaction for the solutes $A$ and $B$, namely
\begin{equation}
a_r A \to B.
\end{equation}
The transport and reaction process is
described by the following advection-diffusion-reaction equations
\begin{equation}
        \left \{
	\begin{aligned}
		\psi\frac{\partial C_A}{\partial t} + q\frac{\partial C_A}{\partial x} &= \psi D\frac{\partial^2 C_A}{\partial x^2}-\psi v_A k_{f,r}C_A^{a_r} , \qquad &(x,t)&\in [0,5]\times[0,1] \\
		\psi\frac{\partial C_B}{\partial t} + q\frac{\partial C_B}{\partial x} &= \psi D\frac{\partial^2 C_B}{\partial x^2}-\psi v_B k_{f,r}C_A^{a_r} , \qquad &(x,t)&\in [0,5]\times[0,1] \\
		C_A(x,0)&=1, C_B(x,0)=0,     &x&\in[0,5] \\
		C_A(0,t)&=1, C_B(0,t)=0,     &t&\in[0,1] \\
		\frac{\partial C_A(5,t)}{\partial x}&=\frac{\partial C_B(5,t)}{\partial x} =0, &t&\in[0,1] \\
	\end{aligned}
        \right .
,
\end{equation}
where $C_A$ and $C_B$ are the concentration of solutes $A$ and $B$ respectively, $q=0.5$ is the Darcy velocity, $\psi=0.4$ is the porosity, $D=1\times 10^{-8}$ is the diffusion coefficient, $v_A=a_r$ and $v_B=-1$ are the stoichiometric coefficients.
The constant $k_{f,r}$ is the chemical reaction rate and $a_r$ is the order of the chemical reaction, both of which are difficult to measure and will be inferred by 
the observations of $C_A$.

To obtain the observations, numerical simulations are conducted using a second-order finite difference method for spatial discretization and the  second-order backward difference method for implicitly marching in time. In simulations,  the mesh size is $\Delta x=0.0125$ and the  time step size is $\Delta t=0.005$. According to \cite{meng2020composite}, we work on inferring the newly defined effective reaction rate $k_f=v_A k_{f,r}$ as well as $a_r$. We set the exact values to $k_f=1.577$ and $a_r=2$. The high-fidelity observations at $(x,t) \in \{0.625, 1.25, 2.5,3.75\} \times \{0.5, 1.0\}$ are extracted from the high-fidelity simulations with the exact $k_f$ and $a_r$ .  As for the low-fidelity observations, they are generated by a low-fidelity simulation with a initial guess of $k_f$ and $a_r$, uniformly sampled in the ranges $[0.75k_f, 1.25k_f]$ and $[0.75a_r, 1.25a_r]$, respectively. The mesh size and time step size in the low-fidelity simulation are the same with those in the high-fidelity simulation. For training networks, only 50 residual points are randomly chosen from the physical domain.

The user-defined hyperparameters for building the multi-fidelity network are as follows. The encoder net and decoder net are built with 6 layers, each layer containing 30 neurons. The feature depth is $L_f=6$, implying the 6th layer is chosen as the feature layer. The feature distance is chosen as $d_f=1.0$. For this problem,  we do not use the Fourier feature embedding.

In network training,  $k_f$ and $a_r$ are initialized by uniformly sampling in the ranges $[0.75k_f, 1.25k_f]$ and $[0.75a_r, 1.25a_r]$, respectively. The inferred values $k_f$ and $a_r$ of the present work and available works in the literature  are given in Table \ref{table_accuracy_InverseReaction}. It is shown that the inferred results are highly accurate with a small relative error and variance. Our results are significantly outperforming the reported results in \cite{meng2020composite}. It is worth noting that only 50 residual points are used for training the networks, while  30000 residual points are used in \cite{meng2020composite}.

\begin{figure}[htbp]
	\centering	
	\includegraphics[width=14.4cm]{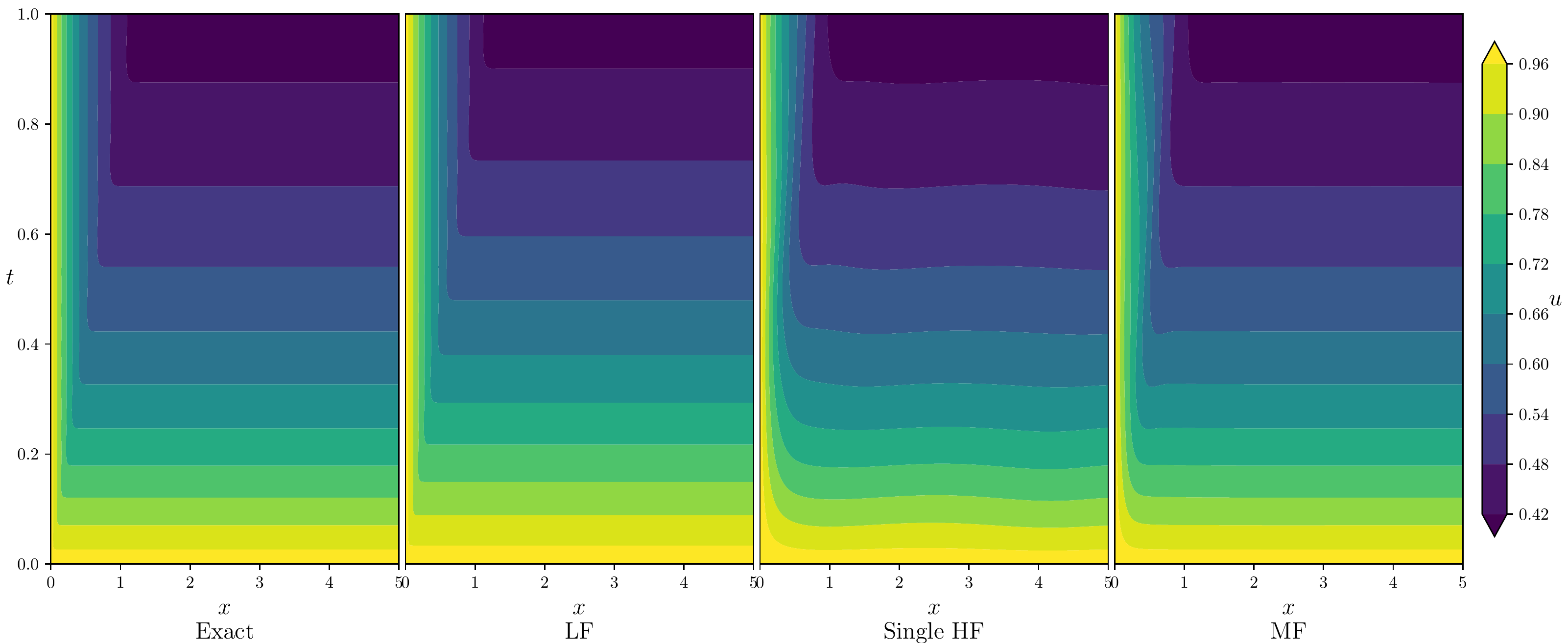}
	\caption{Inverse chemical reaction model: predictions  of the MF approach and the single HF approach versus the reference  high-fidelity (Exact) solution, low-fidelity  solution.}
	\label{fig_contourf_InverseReaction}		
\end{figure}

In order to take a further look at the prediction details, Fig. \ref{fig_contourf_InverseReaction} illustrates the predictions from the MF approach and the single HF approach. The exact solution is featured by the right-angle turns of isolines. The MF approach approximates the right-angle turns more accurately than the single HF approach. A possible explanation is that the MF approach learns from the low-fidelity solution, which is also featured by the right-angle turns of isolines.

\begin{table}[tbp]
	\newcommand{\tabincell}[2]{\begin{tabular}{@{}#1@{}}#2\end{tabular}}
	\centering
	\caption{Summary of  inferred parameters of the single HF approach and the MF approach  for reactive transport. The mean and standard deviation are calculated over 10 independent runs.}
	\begin{tabular}{ccccc}
		\toprule
		Method & $k_f$ &  Error &  $a_r$ & Error \\
		\midrule
		Exact & 1.577 & - & 2  & -\\
		Ref. \cite{meng2020composite} with Single HF & $0.441$ & 72.04\% & $0.558 $  & 72.10\% \\
		Present with Single HF  & $1.583 \pm 1.29\times 10^{-2}$ & 0.39\% & $2.000 \pm 1.41\times 10^{-2}$  & 0.29\% \\
		Ref. \cite{meng2020composite} with MF & $1.414 \pm 7.45\times 10^{-3}$ & 10.33\% & $1.780 \pm 9.44\times 10^{-3}$  & 11.00\% \\
		Present with MF & $1.576 \pm 3.97\times10^{-3}$ & 0.05\% & $2.000 \pm 3.58\times 10^{-3}$ & 0.04\% \\
		\bottomrule
	\end{tabular}
	\label{table_accuracy_InverseReaction}
\end{table}

 %\vspace{-0.5cm}

\subsubsection{Reynolds number discovery for lid-driven flow problem}
In this section, we further consider a realistic application scenario of incompressible flow. The objective is to infer the Reynolds number from observations of the pressure on the boundaries, which are cheap to obtain. The geometry and boundary conditions of the lid-driven flow problem is same with those in Section \ref{subsec_lid_steady}. We assume the exact Reynolds number is $Re=1000$. 

The high-fidelity observations are obtained from numerical simulations, since we have no experimental data. Same with Section \ref{subsec_lid_steady}, we use a resolution of (61, 61) for conducting the high-fidelity simulation. As for the high-fidelity observations, we put $n_s$ equal-spaced probes on each boundary, i.e., there will be $4n_s$ pairs of pressure and positions. The positions are $(x,y)=\{i/(n_s+1)\}_{i=1}^{n_s}\times\{0,1\}$ on the horizontal boundaries and  $(x,y)=\{0,1\}\times\{i/(n_s+1)\}_{i=1}^{n_s}$ on the vertical boundaries. As for the low-fidelity data, they are obtained from either numerical simulations or experiment measurements at different operating conditions.
Here we use the simulations from $Re=400$ with a coarse resolution (11, 11). The low-fidelity data set is collected from the points on a  $51\times 51$ uniform grid  along with the pressure and velocity values interpolated (using Chebyshev spectral expansion) from the low-fidelity simulations. For training networks, 4096 residual points are randomly chosen from the physical domain.

The user-defined hyperparameters for building the multi-fidelity network are as follows. The encoder net and decoder net are built with 6 layers, each layer containing 50 neurons. The feature depth is $L_f=6$, implying the 6th layer is chosen as feature layer. Feature distance is chosen as $d_f=1.0$. The Fourier feature embedding for the input is constructed with
$m=100$ Fourier features and  the average wave number $\pmb{\sigma}=\{\sigma_x, \sigma_y\}=\{0.5,0.5\}$ for $x$ and $y$ directions.

\begin{figure}[htbp]
	\centering	
	\includegraphics[width=14.4cm]{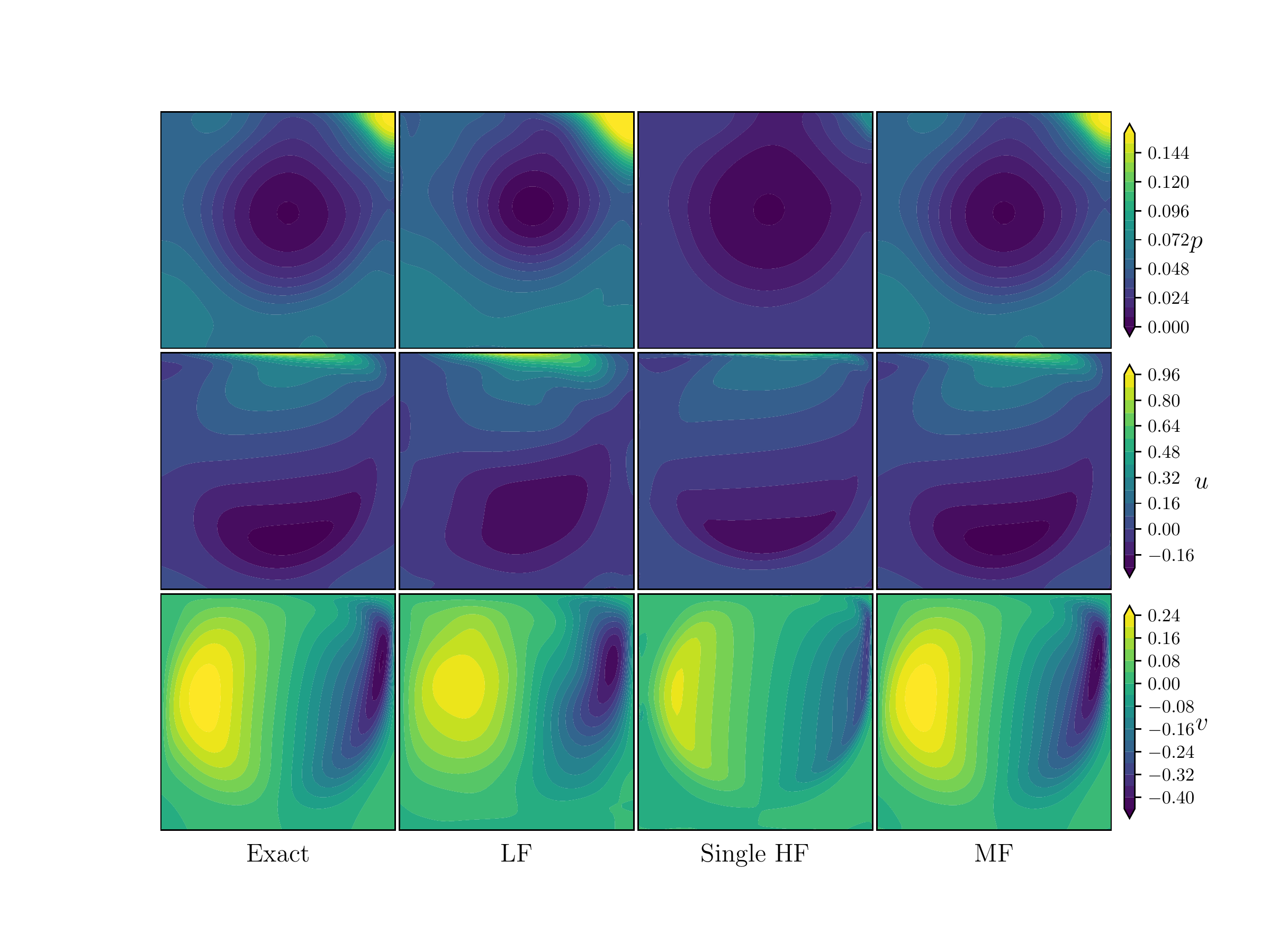}
	\caption{Inverse lid-driven flow: predictions  of the MF approach and the single HF approach versus the reference  high-fidelity (Exact) solution and low-fidelity  solution.}
	\label{fig_contourf_InverseLidDriven}		
\end{figure}

In network training, $Re$ is initialized by randomly sampling in the range [200,5000].
The inferred values of $Re$  is given in Table \ref{table_accuracy_lid_inverse}. Not surprising, the prediction accuracy of both the MF approach and the single HF approach increases with the number of observations. However, the MF approach can reach a much higher accuracy than the single HF approach, when only a small number of observations are available.
Fig. \ref{fig_contourf_InverseLidDriven} illustrates the predictions of flow field from the MF approach and the single HF approach for $n_s$=3. The MF approach predicts the solution accurately with the guidance from the low-fidelity solution. However, the single HF approach fails in resolving the boundary layer of the right sidewall, where the boundary thickness is underestimated. This leads to an overestimation of the Reynolds number, namely $Re=7888$, for the single HF approach. 
When $n_s=3$, the prediction accuracy of the MF approach is $0.40\%$, which is much better than the single HF approach with $n_s=9$.
It indicates that the MF approach can significantly reduce the amount of high-fidelity observations required for inverse discovery by using low-fidelity data, and thus save the cost of generating more high-fidelity observations. 

\begin{table}[tbp]
	\newcommand{\tabincell}[2]{\begin{tabular}{@{}#1@{}}#2\end{tabular}}
	\centering
	\caption{Summary of  inferred Reynolds number of the single HF approach and the MF approach  for lid-driven flow, given $4n_s$ observations of pressure. The mean and standard deviation are calculated over 5 independent runs.}
	\begin{tabular}{cccc}
		\toprule
		Method & $n_s$ &$Re$ &  Error  \\
		\midrule
		Exact & -&1000  & -\\
		\multirow{3}{*}{ Single HF}& 
		 3&$2347.30 \pm 2770$ & 134.73\%  \\
		&9&$986.548 \pm 4.689$ & 1.35\%  \\
		&27&$999.394 \pm 1.052$ & 0.06\%  \\
		\multirow{3}{*}{ MF}& 
		3&$995.972 \pm 1.795$ & 0.40\%  \\
		&9&$999.446 \pm 0.250$ & 0.05\%  \\
		&27&$999.552 \pm 0.181$ & 0.04\%  \\
		\bottomrule
	\end{tabular}
	\label{table_accuracy_lid_inverse}
\end{table}

 %\vspace{-0.5cm}

\FloatBarrier
\section{Discussion and future work}
\label{sec_Discussion}
In this work, we propose a multi-fidelity machine learning architecture for inverse discovery and forward prediction of nonlinear problems. The basic idea behind the architecture is that the discrepancy between the low-fidelity and high-fidelity solutions is modeled by constraining their relative distance in a feature-adjacent space.  With respect to the space-time coordinates,  the function basis of the feature-adjacent space is approximated by an encoder net. The transformation between the feature-adjacent space and the solution space is approximated by a decoder net. The networks are trained with a self-adaptive weighting method. 

The multi-fidelity architecture is first thoroughly studied on the challenging steady lid-driven problem for Reynolds number up to 5000. The relative error is of order $10^{-4}$, which is 2-3 orders of magnitude lower than the low-fidelity solutions. Second, the multi-fidelity architecture is tested by a series of time-dependent problems described by ODEs and PDEs. The long-term prediction capability of the multi-fidelity architecture is studied by increasing the time span to be solved.  The multi-fidelity approach  outperforms the single high-fidelity approach especially for the larger time span. For the ODE and 1D PDE problems, the multi-fidelity approach can achieve a relative error about 2-3 orders lower than that of the low-fidelity solution. For the complex 2D PDE problem,  the multi-fidelity approach can achieve a relative error about 1 order lower than that of the low-fidelity solution.
Third, the multi-fidelity architecture is further tested by three inverse problems. The multi-fidelity approach is much better than the single high-fidelity approach, and is able to significantly reduce the number of high-fidelity observations for parameter discovery while maintaining high accuracy in the inferred parameters. 

The present work proposes a fundamentally new strategy for building the relation between the low- and high-fidelity solutions. The proposed multi-fidelity architecture has the potential to advance our simulation capability in more general scenarios, such as 1) problems with prior low- and high-fidelity observations. 2) problems with prior low-fidelity physics and high-fidelity observations, 3)
the forward  prediction or inverse discovery of parameterized PDEs.

\section*{Acknowledgments}
The authors would like to thank Dr. Amanda Howard for very helpful discussions. The work is supported by the U.S. Department of Energy, Advanced Scientific Computing Research program, under the Physics-Informed Learning Machines for Multiscale and Multiphysics Problems (PhILMs) project (Project No. 72627).
Pacific Northwest National Laboratory (PNNL) is a multi-program national laboratory operated for the U.S. Department of Energy (DOE) by Battelle Memorial Institute under Contract No. DE-AC05-76RL01830.

%\section*{References}
\bibliographystyle{elsarticle-num}
\bibliography{bibliography}

\end{document}